\documentclass[lettersize,journal]{IEEEtran}
\pdfoutput=1
\usepackage[hidelinks=true]{hyperref}
\usepackage{amsmath,amsfonts}
\usepackage{algorithm2e}
\usepackage{array}
\usepackage[caption=false,font=normalsize,labelfont=sf,textfont=sf]{subfig}
\usepackage{textcomp}
\usepackage{stfloats}
\usepackage{verbatim}
\usepackage{orcidlink}
\usepackage{xspace}
\usepackage{booktabs}
\usepackage{xcolor}
\usepackage{bbding}
\usepackage{tcolorbox}
\usepackage{multicol}
\usepackage{multirow}
\usepackage{graphicx}
\usepackage{cite}
\SetKwComment{Comment}{// }{}
\RestyleAlgo{ruled}

\newcommand\edge{\textit{EdgeCompress}\xspace}
\newcommand\ssd{EC-DIC\xspace}
\newcommand\css{EC-Static\xspace}

\newcommand\rcc{RCC\xspace}
\newcommand\sota{SOTA\xspace}

\newcommand\major{\textcolor{black}}
\newcommand\minor{\textcolor{black}}

\begin{document}

\title{\textit{EdgeCompress}: Coupling Multidimensional Model Compression and Dynamic Inference for EdgeAI}

\author{Hao Kong \orcidlink{0000-0002-1378-0056}, Di Liu \orcidlink{0000-0002-4365-2768},~\IEEEmembership{Member,~IEEE,} Shuo Huai \orcidlink{0000-0002-4744-304X}, Xiangzhong Luo \orcidlink{0000-0002-0758-2248}, Ravi Subramaniam \orcidlink{0000-0002-7118-0796}, Christian Makaya, Qian Lin, Weichen Liu \orcidlink{0000-0001-9348-4662},~\IEEEmembership{Member,~IEEE}

\thanks{
\copyright~2023 IEEE. Personal use of this material is permitted. Permission from IEEE must be obtained for all other uses, in any current or future media, including reprinting/republishing this material for advertising or promotional purposes, creating new collective works, for resale or redistribution to servers or lists, or reuse of any copyrighted component of this work in other works. This is the author's accepted version of the article published in IEEE Transactions on Computer-Aided Design of Integrated Circuits and Systems, vol. 42, no. 12, pp. 4657-4670, Dec. 2023, DOI: 10.1109/TCAD.2023.3276938.
Corresponding author: Weichen Liu (email: liu@ntu.edu.sg)

Hao Kong, Shuo Huai are with the School of Computer Science and Engineering, Nanyang Technological University (NTU), Singapore, and also with HP-NTU Digital Manufacturing Corporate Lab, Nanyang Technological University (NTU), Singapore. 

Di Liu is with Department of Computer Science, Norwegian University of Science and Technology (NTNU), Trondheim, Norway. 

Xiangzhong Luo and Weichen Liu are with the School of Computer Science and Engineering, Nanyang Technological University (NTU), Singapore. 

Ravi Subramaniam, Christian Makaya, and Qian Lin are with HP Inc., Palo Alto, CA, USA. 
}
}



\maketitle

\begin{abstract}
\major{Convolutional neural networks (CNNs) have demonstrated encouraging results in image classification tasks.}
However, the prohibitive computational cost of CNNs hinders the deployment of CNNs onto resource-constrained embedded devices. To address this issue, we propose \edge, a comprehensive compression framework to reduce the computational overhead of CNNs.
In \edge, we first introduce dynamic image cropping, where we design a lightweight foreground predictor to accurately crop the most informative foreground object of input images for inference, which avoids redundant computation on background regions. Subsequently, we present compound shrinking to collaboratively compress the three dimensions (depth, width, and resolution) of CNNs according to their contribution to accuracy and model computation. Dynamic image cropping and compound shrinking together constitute a multi-dimensional CNN compression framework, which is able to comprehensively reduce the computational redundancy in both input images and neural network architectures, thereby improving the inference efficiency of CNNs.
Further, we present a dynamic inference framework to efficiently process input images with different recognition difficulties, where we cascade multiple models with different complexities from our compression framework and dynamically adopt different models for different input images, which further compresses the computational redundancy and improves the inference efficiency of CNNs, facilitating the deployment of advanced CNNs onto embedded hardware. Experiments on ImageNet-1K demonstrate that \edge reduces the computation of ResNet-50 by 48.8\% while improving the top-1 accuracy by 0.8\%. Meanwhile, we improve the accuracy by 4.1\% with similar computation compared to HRank. the state-of-the-art compression framework.
The source code and models are available at \textcolor{blue}{\url{https://github.com/ntuliuteam/edge-compress}}
\end{abstract}

\begin{IEEEkeywords}
Embedded systems, neural network compression, hardware/software co-design, dynamic neural network
\end{IEEEkeywords}

 




\section{Introduction}
\label{sec:intro}

\major{Convolutional neural networks (CNNs) have gained popularity in image classification tasks \cite{deng2009imagenet}.}
Benefiting from the advances in high-quality datasets \cite{deng2009imagenet,lin2014microsoft} and network architecture designs \cite{tan2019efficientnet,radosavovic2020designing}, the accuracy of modern CNNs has been constantly improved. Nevertheless, such accuracy improvement comes at higher computational overhead \cite{tan2019efficientnet,radosavovic2020designing,liu2022convnet}. 
Recently, to mitigate data transmission latency and respond to the growing concern about data privacy, a new paradigm named EdgeAI has emerged, \major{which deploys CNNs onto embedded devices near users to process data locally instead of uploading data to the cloud \cite{shi2016edge,Chen_2022_CVPR}}. However, embedded devices are usually resource-constrained and in turns are unable to accommodate resource-hungry CNNs.
To facilitate the deployment of advanced CNNs onto resource-constrained embedded devices, efforts have been made to compress the computational overhead of CNNs.

\begin{figure}
    \centering
    \includegraphics[width=0.45\textwidth]{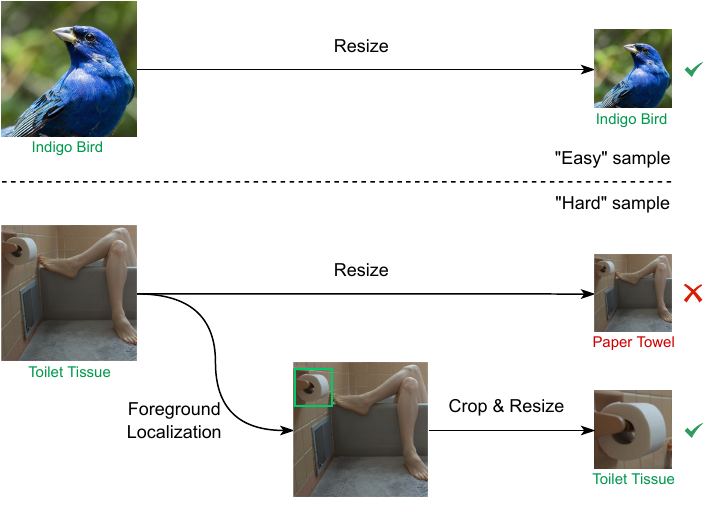}
    \caption{The predictions from ResNet-50. For easy samples, the network can still generate correct predictions at a smaller resolution (e.g., 112 $\times$ 112 for ImageNet-1K). For hard samples, simply resizing images to a smaller resolution can lead to misclassification, while using dynamic cropping can correctly classify hard samples at a smaller resolution.}
    \label{fig:motivation}
\end{figure}

The computation of a CNN, i.e., the Multiply-Accumulate Operations (MACs), mainly results from two aspects: 1) high-resolution input images and 2) gigantic network architectures. To reduce the computational redundancy in input images (i.e., spatial redundancy), 
\major{many works propose to reduce the resolution (i.e., the height or width of input images) for inference \cite{sandler2018mobilenetv2, tan2019mnasnet, zhu2021dynamic}.}
However, as shown in the motivational example in Fig. \ref{fig:motivation}, this coarse spatial redundancy reduction approach is only effective for images with clear foreground. For images in which the foreground only occupies a small portion of the whole image, directly shrinking the whole image will lose important features of the foreground, leading to a wrong prediction. This observation triggers our first motivation: 
\begin{tcolorbox}
\textbf{Motivation 1:} Can we reduce the inference resolution of input images without sacrificing accuracy?
\end{tcolorbox}

\major{On the other hand, network pruning \cite{lin2020hrank,wang2021convolutional,yu2019slimmable,molchanov2019importance,wang2019dbp,alwani2022decore} is also proposed to compress the depth (i.e., the number of layers) and width (i.e., the number of channels in each layer) of the network architecture.
Specifically, to optimize the efficiency of CNNs, width pruning \cite{lin2020hrank,molchanov2019importance,wang2021convolutional,alwani2022decore} devotes to removing less sensitive channels in each layer to yield `thinner' networks, while depth pruning \cite{wang2019dbp} conducts pruning at a coarser granularity (i.e., layer), which directly removes unimportant layers to construct `shallower' networks.}
\major{However, the above techniques only reduce the redundancy in a single dimension of CNNs while ignoring the redundancy in the other dimensions.}
Such single-dimensional compression approaches can only achieve a very limited compression rate. This phenomenon brings us to the second motivation:
\begin{tcolorbox}
\textbf{Motivation 2:} Can we combine the compression of all three dimensions of CNNs to achieve a higher compression rate while maintaining high accuracy?
\end{tcolorbox}

In addition, given a resource budget, existing approaches usually yield a fixed compressed neural network and resolution for all images. However, as discussed in \cite{zhu2021dynamic, wang2020glance}, different images are of distinct recognition difficulties, using a static model and resolution to process all images can lead to inefficient utilization of computation, achieving only sub-optimal efficiency and accuracy. Practically, for images with simple features, a small CNN model is adequate to generate correct results. For complex images, a larger model with higher capability should be used to extract high-level features for a correct prediction. This inspires our last motivation:
\begin{tcolorbox}
\textbf{Motivation 3:} Can we dynamically adjust the model and resolution for different images during inference to further optimize inference efficiency and accuracy?
\end{tcolorbox}



\major{To address the above questions for more efficient image classification with CNNs,
we, in this paper, propose a novel inference framework, \edge, to comprehensively reduce the inference overhead of CNNs, thereby optimizing the classification efficiency of CNNs and facilitating the deployment of advanced CNNs onto edge devices.} 
In \edge, we first propose a two-stage multi-dimensional model compression framework to coordinately compress all three dimensions of CNNs. 
In the first stage, we introduce a novel dynamic image cropping (DIC) strategy to accurately remove the spatial redundancy in input images, in which we design a lightweight foreground predictor to efficiently localize the most discriminative foreground of input images, \major{then only the detected foreground will be preserved for classification and the redundant background will be discarded. As shown in Figure \ref{fig:motivation}, through the dynamic image cropping strategy, we are capable of generating fine-cropped images with less spatial redundancy, thereby achieving satisfactory classification accuracy even at a smaller resolution.} In the second stage, we present a compound shrinking (CS) strategy to jointly compress the three dimensions of CNNs, thereby further reducing the redundancy in input images and network architectures. We first quantify the impact of shrinking different dimensions on model complexity and accuracy, according to which we automatically calculate a shrinking coefficient for each dimension to coordinate the shrinking of different dimensions to achieve a higher compression rate while still maintaining the accuracy. By the means of the two-stage multi-dimensional compression framework, given a computation budget, we are able to comprehensively reduce redundant computation to meet the budget without sacrificing accuracy obviously.
\major{Based on the compression framework, we further propose a novel dynamic inference framework to adaptively process different input images with different models and resolutions at runtime. First, we utilize the compound shrinking strategy to compress the give baseline network and generate multiple sub-networks with diverse model sizes and accuracy, which are then cascaded in ascending order of the model size and then each input image will be processed by those models sequentially.} At the end of the inference of each model, we propose a novel metric to evaluate the confidence of the prediction result. Once a confident prediction is obtained, the dynamic inference will be terminated without executing subsequent models. In practice, most input images can be confidently recognized by early models with small computational overhead, while large models will be activated only for a few hard samples. Consequently, compared to static inference with a single model, the overall computational complexity of our dynamic inference is reduced significantly without compromising accuracy.
Our main contributions are summarized as follows:
\begin{enumerate}
    \item We propose dynamic image cropping to reduce the spatial redundancy in images, where we design a lightweight detector to efficiently localize the foreground area of an image and conduct instance-aware dynamic cropping. Those finely cropped images can be correctly recognized even at a smaller resolution, which greatly reduces the computational cost of CNNs.
    
    
    \item We also propose compound shrinking to jointly compress the three dimensions of a CNN. We first quantify the impact of each dimension on accuracy and model complexity, and then generate the optimal joint compression strategy accordingly. By this means, we greatly reduce the redundancy in both input images and network architectures for a higher compression rate.
    
    \item We further introduce a dynamic inference framework to efficiently process input images with different recognition difficulties. We cascade multiple models from our compression framework and adaptively utilize different models and resolutions for different images. In this way, we effectively adjust the computational cost for different input images, reducing the overall computational cost without compromising the final accuracy.

    \item We seamlessly integrate the dynamic image cropping, compound shrinking, and dynamic inference into a deep compression framework (i.e., \edge) for efficient deep learning inference, which can optimally adapt the model cost to meet different resource constraints of embedded hardware while maximizing model accuracy.
    
\end{enumerate}

Extensive experiments demonstrate the advantages of the proposed \edge over other SOTA model compression approaches. Specifically, \edge reduces the MACs of ResNet-50 by 48.8\% while improving the top-1 accuracy by 0.8\% on ImageNet-1K. Moreover, compared to the SOTA compression framework, HRank \cite{lin2020hrank}, \edge also achieves 4.1\% higher accuracy with similar model MACs.


\section{Related Work}
\label{sec:related}
\edge makes innovative contributions mainly in three areas: 1) object localization, 2) CNN compression, and 3) dynamic neural networks. Therefore, we discuss the related works in this section.

\begin{figure*}
    \centering
    \includegraphics[width=0.95\textwidth]{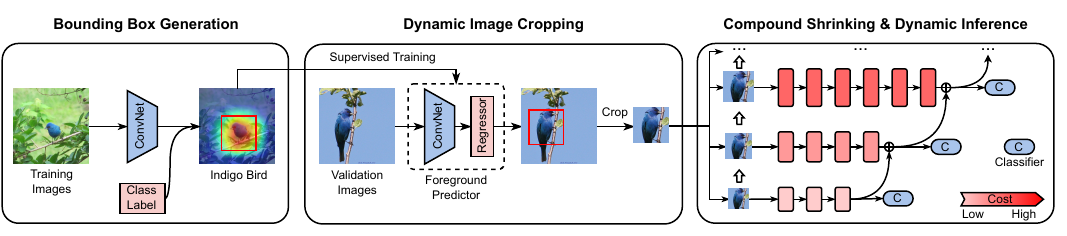}
    \caption{The overview of the proposed \edge framework, which mainly consists of four components: bounding box generation (BBG), dynamic image cropping (DIC), compound shrinking (CS), and dynamic inference (DI).}
    \label{fig:framework}
\end{figure*}

\noindent
\textbf{Object localization: }
Object localization algorithms focus on efficiently detecting the location of foreground objects in an image, which can be mainly divided into supervised object localization (SOL) and weakly supervised object localization (WSOL). SOL \cite{bochkovskiy2020yolov4,liu2016ssd,ren2015faster,he2017mask} achieves promising accuracy with the help of well annotated datasets like COCO \cite{lin2014microsoft} and Pascal VOC \cite{Everingham10}. However, the difficulties in building larger detection datasets hinder the further development of SOL. WSOL \cite{wei2019unsupervised,zhang2020rethinking,zhou2016learning,selvaraju2017grad,zhang2018adversarial} can coarsely localize objects of interest with only image-level labels, which makes WSOL applicable to more large-scale datasets without position annotations, such as ImageNet-1K \major{(also known as ImageNet or ILSVRC-2012)} \cite{deng2009imagenet}. Specifically, CAM \cite{zhou2016learning} and Grad-CAM \cite{selvaraju2017grad} utilize a well-trained CNN to quantify the importance of each pixel of an image and determine the position of the most contributing part accordingly, i.e., the foreground. Further, ACoL \cite{zhang2018adversarial} presents a novel CNN with two branches to adversarially learn the full region of objects, improving the localization accuracy. 
However, the huge computational cost and latency make both the SOL (e.g., SSD) and WSOL inapplicable in our context, i.e., resource-constrained embedded hardware.

\noindent
\textbf{CNN compression: }
Computational redundancy widely exists in CNNs \cite{liu2022brining}. To achieve a better trade-off between model accuracy and execution efficiency, effort have been made to reduce the redundancy from different dimensions of CNNs. Specifically, depth pruning \cite{wang2019dbp,zhang2022layer} devotes to compressing layer-level redundancy, which removes the entire layer with low sensitivity. Channel pruning \cite{lin2020hrank,molchanov2019importance,sandler2018mobilenetv2,yu2019slimmable,alwani2022decore} conducts pruning at a finer granularity, which builds compact CNNs by removing unimportant channels from each layer, which reduces the computation and memory footprint of CNNs. Among channel pruning approaches, MobileNetV2 \cite{sandler2018mobilenetv2} and Slimmable networks \cite{yu2019slimmable} remove channels from all layers uniformly, while Taylor pruning \cite{molchanov2019importance}, HRank \cite{lin2020hrank}, and DECORE \cite{alwani2022decore} evaluate the global importance of each channel and then prune channels in a layer-wise manner. Both depth pruning and channel pruning focus on compressing the network architecture, while resolution pruning \cite{sandler2018mobilenetv2,tan2019mnasnet,zhu2021dynamic,wang2020glance} optimizes the spatial redundancy in input images by shrinking images to smaller resolutions or selectively cropping images for inference. MNasNet \cite{tan2019mnasnet} reduces the spatial redundancy by utilizing a fixed small resolution for all images during inference. Instead, DR-ResNet \cite{zhu2021dynamic} introduces a dynamic resolution strategy to dynamically assign different resolutions to different images according to their recognition complexity. Moreover, GFNet \cite{wang2020glance} introduces a Glance-and-Focus inference strategy, which utilizes both the shrunk version and small local patches of an image for inference to accelerate CNNs while preserving high accuracy.
However, all above methods only consider reducing the redundancy in a single dimension and thus only achieve a limited compression rate. In contrast, jointly compressing all dimensions promises a better trade-off between the compression rate and accuracy.



\noindent
\textbf{Dynamic neural networks: }
To efficiently process input images corresponding to diverse classification difficulties and reduce the computational redundancy, dynamic inference approaches are proposed to adaptively utilize different models or different parts of a model for different images. Early-exit networks \cite{teerapittayanon2016branchynet, huang2018multiscale, bolukbasi2017adaptive} insert multiple intermediate classifiers inside the network and allow easy samples to exit at shallow layers without executing deeper layers. Different from Early-exit networks that execute layers densely, Layer-skipping networks \cite{graves2016adaptive, wang2018skipnet, veit2018convolutional} selectively skip less important intermediate layers to avoid redundant computation and optimize the execution efficiency. Channel-skipping networks \cite{hua2019channel,lin2017runtime,gao2018dynamic,herrmann2020channel} consider reducing redundant computation in the width dimension, which introduce channel gates to control the execution of each channel and different channels will be selectively activated for different samples. More recently, instead of dynamically changing the network architecture at runtime, resolution-level dynamic inference approaches \cite{gao2018dynamic, yang2020resolution} are proposed to dynamically adjust the resolution of input images during inference. Specifically, easy samples are allowed to inference at a small resolution, which greatly optimizes the inference cost. In spite of the efficiency improvement achieved by above methods, they only focus on adjusting a single dimension at runtime, which can only achieve sub-optimal efficiency and accuracy. Instead, our dynamic inference framework utilizes the models generated from our multi-dimensional compression framework, which enables multi-dimensional dynamic inference and thus achieves higher accuracy and efficiency. 



\section{The Proposed EdgeCompress Framework}
\label{sec:method}

In this section, we first outline the design of \edge and then describe each component in detail.

As demonstrated in Fig. \ref{fig:framework}, 
\major{before inference, we first utilize Grad-CAM to generate the salience map of all training images in the classification dataset $\mathcal{D}$, and then we generate a bounding box for each image according to the salience map and form a pseudo bounding box label set $\mathcal{B}$. Thereafter, we exploit the image-box pairs \{$\mathcal{D}_i$, $\mathcal{B}_i$\} to train a lightweight predictor.} Meanwhile, we use compound shrinking to jointly compress the three dimensions of a CNN and generate multiple CNNs with different computational complexities, which are then cascaded for dynamic inference.
\major{In inference, the input image will be first fed into the trained predictor to efficiently localize the foreground object. Thereafter, the foreground object will be cropped and sequentially sent to the CNN models generated by compound shrinking for dynamic inference. Once a confident prediction is obtained, the inference will be terminated immediately without executing subsequent models.}

\subsection{Bounding Box Generation}
\label{subsec:generate}


\major{As aforementioned, dynamically cropping the foreground for inference is promising in reducing computation and improving classification accuracy.} However, for classification datasets like ImageNet-1K, there is no out-of-the-box position annotation for the foreground object. Moreover, the position of the foreground object varies in different images, which makes it difficult to efficiently localize the foreground object.

\begin{figure}
    \centering
    \includegraphics[width=0.45\textwidth]{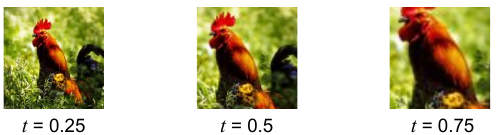}
    \caption{\major{By applying different salience threshold $t$, we can obtain different cropped images. The larger the threshold value, the more radical the cropping.}}
    \label{fig:threshold}
\end{figure}


To address this limitation, we first use Grad-CAM \cite{selvaraju2017grad} to automatically generate the position annotations. 
Specifically, let the class label of the given image be $c$. We first perform forward inference with a well-trained CNN (e.g. ResNet-50) to obtain the prediction score $p^c$ for class $c$, and then conduct backpropagation to compute the gradient of the score $p^c$ with respect to each activation of the last convolutional layer. Thereafter, the gradients are aggregated within each channel via global average pooling. The obtained scalar for each channel can be seen as the weight of the channel, which can be calculated as follows:
\begin{equation}
\small
    a_k^c = \overbrace{\frac{1}{Z}\sum_i\sum_j}^{\textit{pooling}} \overbrace{\frac{\partial p^c}{A_{ij}^k}}^{\textit{gradients}}
\end{equation}
where $a_k^c$ is the weight of channel $k$ for class $c$, and $A_{ij}^k$ is a single activation indexed by $i$ and $j$ in the 2-D feature map of channel $k$. With the weights of all channels determined, the salience map for class $c$ can be obtained by computing the weighted sum of all feature maps over the channel dimension, which is formulated as:
\begin{equation}
\small
    L_{Grad\_CAM}^c = ReLU\underbrace{\left(\sum_k a_k^c A^k\right)}_{\textit{linear combination}}
\end{equation}
where $A^k$ is the 2-D feature map of channel $k$, and ReLU is used to eliminate the impact of negative activations. Finally, the obtained salience map is upsampled to the same size as the input image via bi-linear interpolation algorithm.

With the salience map generated, we then introduce a simple yet effective strategy to determine the bounding box of the foreground object. Initially, we set the box as the boundary of the image. Subsequently, we shrink the four sides of the box simultaneously, and once a side reaches our preset salience threshold $t$, the side is frozen. The bounding box is determined after all sides are frozen. Note that it is crucial for the final result to appropriately select the value of $t$. As demonstrated in Fig. \ref{fig:threshold}, a too small threshold will result in residual background redundancy, while a too large threshold will lose some important features. Therefore, we conduct empirical experiments to determine the optimal threshold value. As shown in Table \ref{tab:threshold}, we achieve the highest accuracy when the threshold $t$ is set to 0.5. Therefore, we set $t=0.5$ in our experiments. Note that more fine-grained searching for $t$ may further improve the accuracy, but it also increases the search cost. Finally, the generated box annotations are saved in the form of [$X_{min}$, $Y_{min}$, $X_{max}$, $Y_{max}$], which denotes the boundary of the foreground in the image.

\begin{table}[tbp]
  \centering
  \caption{The impact of using different salience thresholds on prediction accuracy. The model is trained and evaluated on ImageNet-1K. $t=0$ means using the original images without Grad-CAM cropping.}
    \resizebox{0.47\textwidth}{!}{
    \begin{tabular}{lcccr}
    \toprule
    \toprule
    \textbf{Model} & \textbf{\#Params (M)} & \textbf{\#MACs (B)} & \textbf{\textit{t}} & \textbf{Top-1 Acc. (\%)} \\
    \midrule
    \multirow{4}[2]{*}{ResNet-50} & \multirow{4}[2]{*}{25.6} & \multirow{4}[2]{*}{4.1} & 0.00 (Baseline) & 76.02 \\
        & &  & 0.25  & 76.45 \\
        & &  & 0.50   & \textbf{76.88} \\
        & &  & 0.75  & 76.32 \\
    \bottomrule
    \bottomrule
    \end{tabular}%
    }
  \label{tab:threshold}%
\end{table}%

\subsection{Dynamic Image Cropping}
\label{subsec:predict}

\begin{figure}
    \centering
    \includegraphics[width=0.47\textwidth]{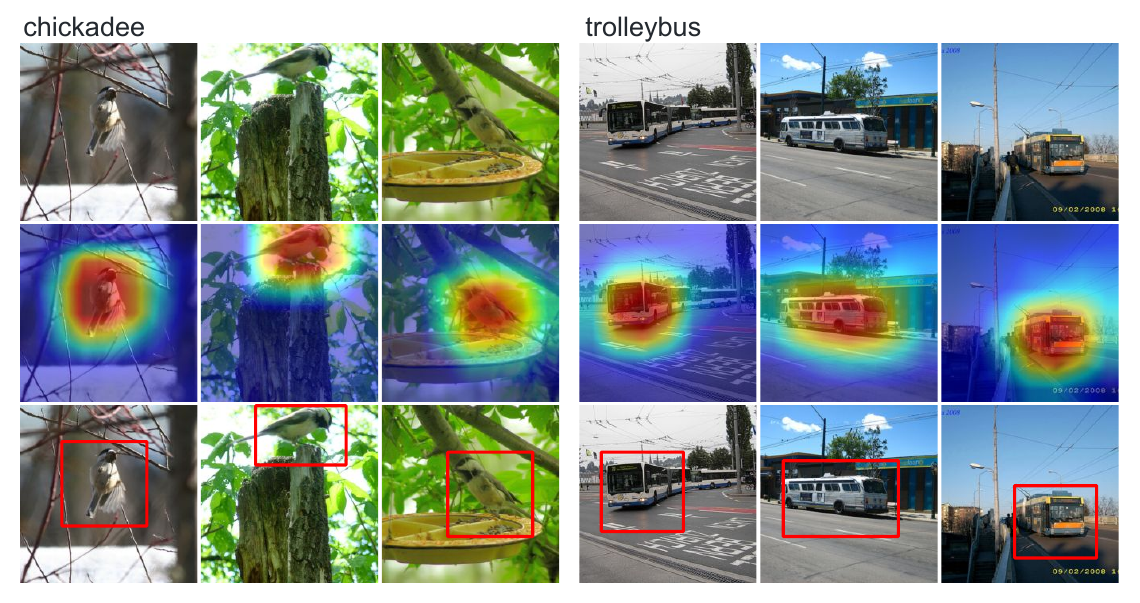}
    \caption{The bounding boxes generated with the salience threshold $t=0.5$, which accurately localize the key object in each image.}
    \label{fig:grad-cam}
\end{figure}

Fig. \ref{fig:grad-cam} shows that we are capable of accurately localizing the foreground of images with Grad-CAM. However, Grad-CAM cannot be directly applied to edge applications because of the time-consuming backpropagation process. Moreover, Grad-CAM requires the class label as weak supervision, which is unavailable for validation images. To address these issues, we design a foreground predictor to efficiently localize the foreground of input images.

\subsubsection{Predictor Architecture}

\major{Existing detection models, such as Faster R-CNN \cite{ren2015faster}, are mainly proposed for object detection tasks (e.g., MS COCO \cite{lin2014microsoft}), which usually contain a large number of parameters and computation to accurately localize and identify the multiple objects in each input image. However, we focus on classification tasks, where each input image contains only one object and thus the localization difficulty is much lower than in detection tasks. Moreover, to achieve dynamic cropping, we only need to output the position of the foreground without predicting its label.  Consequently, existing detection models become redundant and inefficient in our context. To this end, we design a novel lightweight foreground predictor to efficiently localize the unique foreground object of each input image. The details of the proposed foreground predictor is summarized in Table \ref{tab:arch}, which consists of several residual bottleneck blocks \cite{he2016deep} and a fully connected layer.} A residual bottleneck contains two convolutional layers with 1$\times$1 kernels and one convolutional layer with 3$\times$3 kernels in the middle. The computational cost mainly results from the 3$\times$3 convolutional layer. Therefore, to reduce the cost and accelerate the predictor, we only stack two residual bottleneck blocks in each stage and each block is only equipped with a small number of channels. Consequently, the proposed predictor only contains 0.27M parameters and 0.09B MACs, which is negligible compared to popular object detectors (e.g., Faster R-CNN with 134.7M (499$\times$) parameters and 15.1B (167.8$\times$) MACs \cite{li2018tiny}).

\subsubsection{Training of Foreground Predictor}
We train the predictor in a supervised manner. First, we generate a bounding box label set $\mathcal{B}$ for all training images as described in Subsection \ref{subsec:generate}, then the labels are utilized to train the predictor. We use the mean square error (MSE) as the loss function. Let $\mathcal{P}_i=$ [$X_{min}^p$, $Y_{min}^p$, $X_{max}^p$, $Y_{max}^p$] be the output of the predictor, and $\mathcal{G}_i=$ [$X_{min}^g$, $Y_{min}^g$, $X_{max}^g$, $Y_{max}^g$] be the generated box label, the loss function can be formulated as:
\begin{equation}
\footnotesize
\begin{aligned}
    \mathcal{L}_{box}  = & \ MSELoss(\mathcal{P}_i, \ \mathcal{G}_i) \\
                       = & \ \frac{1}{4} ((X_{min}^g-X_{min}^p)^2+(Y_{min}^g-Y_{min}^p)^2 \\
                                & +(X_{max}^g-X_{max}^p)^2+(Y_{max}^g-Y_{max}^p)^2)
\end{aligned}
\end{equation}

To balance the training overhead and prediction accuracy, we train the predictor with Adam \cite{kingma2015adam} optimizer for 40 epochs. The initial learning rate is set to 1e-3, and the learning rate is scheduled using exponential decay \cite{Li2020An}. The training of the box predictor is decoupled with backbone networks. Once the predictor is trained, it can be directly applied to different classification backbones without any training overhead. During inference, the trained predictor will quickly localize the foreground object of the input image and generate a finely cropped image, which significantly reduces the redundancy in the input image.

\subsection{Compound Shrinking}
\label{subsec:compound}

The proposed DIC significantly reduces the redundancy in images, improving the computational efficiency. We observe that redundancy also exists in network architectures (e.g., redundant parameters), and only removing the redundancy in images loses the opportunity to further compress the model for embedded hardware. Besides, \cite{tan2019efficientnet} demonstrates that jointly adjusting different dimensions promises higher accuracy. To this end, we propose a compound shrinking (CS) strategy to jointly compress the three dimensions (depth, width, resolution) of CNNs to further reduce the redundancy in images as well as networks while maintaining the accuracy.


\begin{table}[tbp]
  \centering
  \caption{The architecture of the proposed box predictor. \#C denotes the number of channels and \#L denotes the number of layers.}
    \begin{tabular}{lllll}
    \toprule
    \toprule
    \multicolumn{1}{c||}{\textbf{Stage}} & \multicolumn{1}{c||}{\textbf{Block}} & \multicolumn{1}{c||}{\textbf{Resolution}} & \multicolumn{1}{c||}{\textbf{\#C}} & \multicolumn{1}{c}{\textbf{\#L}} \\
    \midrule
    \multicolumn{1}{c||}{1} & \multicolumn{1}{c||}{Conv 3$\times$3} & \multicolumn{1}{c||}{224 $\times$ 224} & \multicolumn{1}{c||}{16} & \multicolumn{1}{c}{1} \\
    \multicolumn{1}{c||}{2} & \multicolumn{1}{c||}{Residual Bottleneck} & \multicolumn{1}{c||}{112 $\times$ 112} & \multicolumn{1}{c||}{16} & \multicolumn{1}{c}{2} \\
    \multicolumn{1}{c||}{3} & \multicolumn{1}{c||}{Residual Bottleneck} & \multicolumn{1}{c||}{56 $\times$ 56} & \multicolumn{1}{c||}{32} & \multicolumn{1}{c}{2} \\
    \multicolumn{1}{c||}{4} & \multicolumn{1}{c||}{Residual Bottleneck} & \multicolumn{1}{c||}{28 $\times$ 28} & \multicolumn{1}{c||}{32} & \multicolumn{1}{c}{2} \\
    \multicolumn{1}{c||}{5} & \multicolumn{1}{c||}{Residual Bottleneck} & \multicolumn{1}{c||}{14 $\times$ 14} & \multicolumn{1}{c||}{64} & \multicolumn{1}{c}{2} \\
    \multicolumn{1}{c||}{6} & \multicolumn{1}{c||}{Pooling \& Linear} & \multicolumn{1}{c||}{7 $\times$ 7} & \multicolumn{1}{c||}{4} & \multicolumn{1}{c}{1} \\
    \midrule
    \multicolumn{5}{l}{\#Params: 0.27M} \\
    \multicolumn{5}{l}{\#MACs: 0.09B} \\
    \bottomrule
    \bottomrule
    \end{tabular}%
  \label{tab:arch}%
\end{table}%

Intuitively, shrinking different dimensions has different impacts on accuracy and model overhead. The core of our compound shrinking strategy is to calculate a shrinking coefficient for each dimension according to their trade-off between accuracy and model overhead. A larger coefficient denotes more radical shrinking. More specifically, the dimension with a steep accuracy drop during shrinking will be assigned a small shrinking coefficient to prevent severe accuracy degradation. To calculate the shrinking coefficients, we first quantify the trade-off of each dimension between accuracy and model overhead. Here we use MACs as the metric to measure the cost of models, because all three dimensions are related to the MACs of a model while only the depth and width can affect the model parameters. Given a MACs budget $\mathcal{M}$, we first obtain the accuracy drops resulting from separately shrinking different dimensions, which can be represented as:
\begin{equation}
\small
\label{eq:acc}
    \Delta A_s(\mathcal{M}) = A_0 - A_s(\mathcal{M})
\end{equation}
where $s\in\{d, w, r\}$ represents the shrunk dimension, $A_s(\mathcal{M})$ denotes the accuracy of the shrunk model, and $A_0$ is the accuracy of the original model. To comply with the rule that the steeper the drop in accuracy, the smaller the coefficient of the corresponding dimension, we design the following equation to determine the shrinking coefficient for each dimension:
\begin{equation}
\small
\label{eq:coeff}
    \mathcal{C}_s(\mathcal{M}) = \frac{\sqrt[3]{\Delta A_d(\mathcal{M}) \cdot \Delta A_w(\mathcal{M}) \cdot \Delta A_r(\mathcal{M})}}{\Delta A_s(\mathcal{M})}
\end{equation}
where $\mathcal{C}_s(\mathcal{M})$ denotes the shrinking coefficient of the dimension $s$ ($s\in\{d, w, r\}$). Through Equation \ref{eq:acc} and Equation \ref{eq:coeff}, we are able to efficiently calculate the coefficients once we obtain the accuracy degradation of the three dimensions in the given MACs regime.

\begin{figure}
    \centering
    \includegraphics[width=0.48\textwidth]{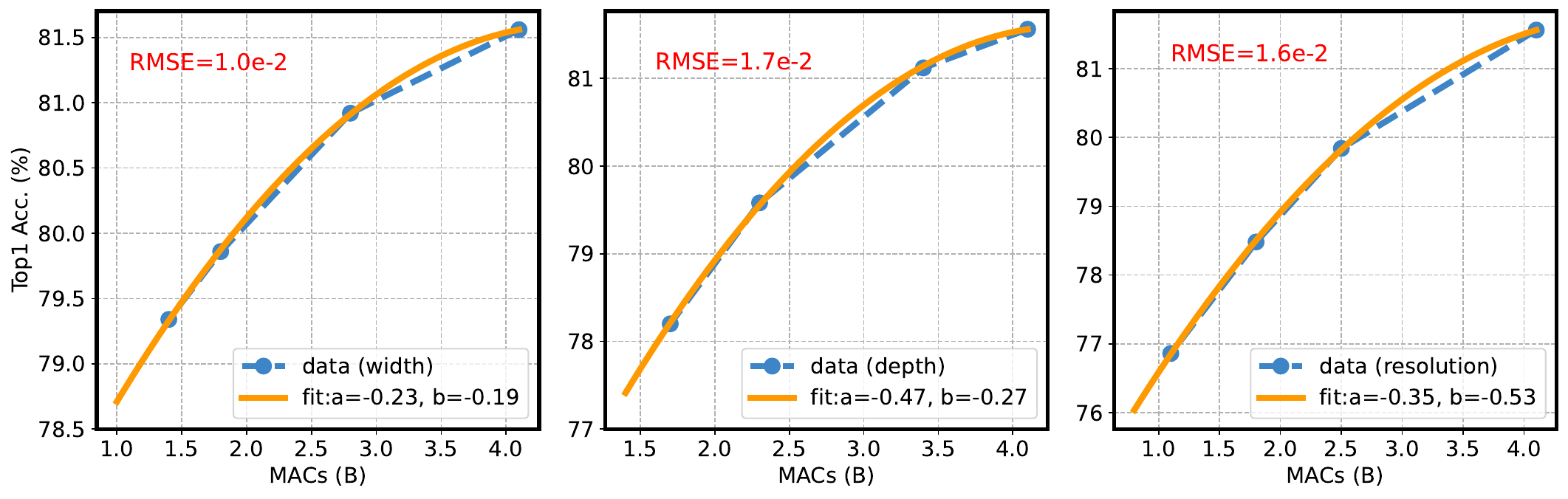}
    \caption{The actual accuracy (blue dotted line) and the estimated accuracy (yellow line) over MACs by separately shrinking the three dimensions. The low root mean square error (RMSE) indicates that the accuracy estimator can well fit the sampled data.}
    \label{fig:dimensions}
\end{figure}

However, the training cost of the compressed models to calculate the accuracy drop is still non-negligible. To mitigate the training overhead, we propose a dimension-wise accuracy estimator to quickly estimate the accuracy of the compressed models and calculate the accuracy degradation resulting from shrinking different dimensions in the given MACs regime. First, we sample a couple of models with different MACs by separately shrinking the three dimensions. As demonstrated in Fig. \ref{fig:dimensions}, the accuracy distribution of the three dimensions along MACs can be well fitted by a quadratic polynomial. Therefore, we design a simple yet effective polynomial estimator to predict the accuracy with respect to the target MACs $\mathcal{M}$. The estimator is formulated as follows:
\begin{equation}
\small
    A_s(\mathcal{M}) = a_s (\mathcal{M} - \mathcal{M}_0)^2 + b_s (\mathcal{M} - \mathcal{M}_0) + A_0 
\end{equation}
where $\mathcal{M}_0$ is the MACs of the original model. $a_s$ and $b_s$ are the hyperparameters to fit for dimension $s$ ($s\in\{d, w, r\}$). Subsequently, we train the dimension-wise estimator using least square regression with the aforementioned sampled data. Fig. \ref{fig:dimensions} shows that the proposed estimator can well fit existing data. Due to the simple and intuitive design of the estimator, we only need to sample and train very few models to train the estimator, and this cost is a one-time cost. \major{With the accuracy estimator established, we are able to quickly estimate the accuracy drop and then calculate the optimal shrinking coefficients for the three dimensions under any given resource constraint. According to the coefficients, we will jointly compress the three dimensions of the baseline network and generate a compact model with optimized efficiency. As the compressed model can be viewed as a subset of the baseline network, we call the compressed model a sub-network.}



\subsection{Dynamic Inference}
\label{subsec:inference}

Through dynamic image cropping and compound shrinking, we can optimally compress a CNN model to different complexities to satisfy various resource constraints in edge environments. Given an embedded device, an intuitive deployment strategy is to select a single model that best fits the hardware capabilities (e.g., memory capacity, computing power) to balance the trade-off between accuracy and execution efficiency. However, as different images correspond to distinct recognition difficulties \cite{zhu2021dynamic, yang2020resolution}, using a single model for all images may over-process simple images and waste resources, while for complex images, the model may under-process them and generate wrong predictions, leading to a sub-optimal trade-off between accuracy and efficiency.

\begin{figure}
    \centering
    \includegraphics[width=0.47\textwidth]{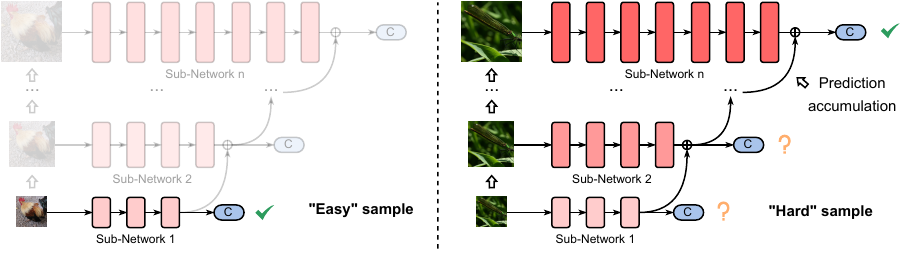}
    \caption{\major{The proposed dynamic inference framework, which utilizes multiple sub-networks to achieve instance-aware inference. These sub-networks are obtained by compressing the baseline network using compound shrinking.}}
    \label{fig:inference}
\end{figure}


\begin{table}[tbp]
  \centering
  \caption{The specifications of the sub-networks generated by the compound shrinking strategy. The baseline network is ResNet-50. The accuracy is measured on ImageNet-1K and the latency is measured on Jetson Nano.}
    \resizebox{0.47\textwidth}{!}{
    \begin{tabular}{crrrr}
    \toprule
    \toprule
    \multicolumn{1}{l}{\textbf{Sub-network Id}} & \multicolumn{1}{l}{\textbf{\#Params (M)}} & \multicolumn{1}{l}{\textbf{\#MACs (B)}} & \multicolumn{1}{l}{\textbf{Latency (ms)}} & \multicolumn{1}{l}{\textbf{Top1 Acc. (\%)}} \\
    \midrule
    1     & 11.65 & 1.21  & 27.15 & 73.86 \\
    2     & 11.79 & 1.30  & 28.05 & 74.21 \\
    3     & 13.53 & 1.84  & 41.62 & 75.56 \\
    4     & 15.40 & 2.40  & 49.41 & 76.32 \\
    5     & 20.30 & 3.22  & 56.01 & 76.81 \\
    6     & 25.90 & 4.20  & 57.09 & 77.20 \\
    \bottomrule
    \bottomrule
    \end{tabular}%
    }
  \label{tab:sub-resnet}%
\end{table}%

\major{To address this problem, we propose a dynamic inference strategy to further optimize the run-time efficiency of CNNs on embedded devices without sacrificing accuracy. As demonstrated in Fig. \ref{fig:inference}, we first apply different MACs constraints to the compound shrinking strategy to generate multiple sub-networks with different accuracy and overhead. The specifications of all sub-networks are summarized in TABLE \ref{tab:sub-resnet}. Thereafter, we deploy the generated sub-networks onto the target hardware before inference and dynamically activate different sub-networks at runtime for better accuracy and efficiency. For easy samples with a distinct foreground, maybe only the smallest sub-network will be activated to efficiently generate the correct prediction, while for hard samples with which small models are unable to produce a confident prediction, larger sub-networks will be gradually activated until a confident prediction is obtained. By doing so, we can avoid unnecessary computation and resource consumption for simple images, improving inference efficiency.} 

\subsubsection{Termination Condition}
Modern large-scale datasets for image classification usually contain millions of images. For example, ImageNet-1K has about 1.3 million images. It is non-trivial to determine when to terminate the inference for each image. Current dynamic inference approaches, such as multi-scale inference \cite{huang2018multiscale} and early-exit networks \cite{laskaridis2020hapi,kaya2019shallow}, exploit the highest prediction probability among all classes as the prediction confidence to control the termination of dynamic inference. Given a CNN $\mathcal{N}$ and an image $x$, the prediction confidence of existing methods can be represented as: 

\begin{equation}
\small
    \label{eq:confidence}
    \begin{aligned}
        \mathcal{I}_{\mathcal{N}}=&\max \left(Softmax(\mathcal{N}(x))\right)  \\
        = &\max\left( \frac{e^{z_{i}}}{\sum_{j=1}^K e^{z_{j}}}\right) \ \ \ for\ i=1,2,\dots,K
    \end{aligned}
\end{equation}
where $z_i$ denotes the $i$-th logit (i.e., the $i$-th output of the fully connected layer) of $\mathcal{N}$, which is transformed into the prediction probability for the $i$-th class with the Softmax function. With Equation \ref{eq:confidence}, the prediction confidence can be efficiently calculated using the output of the network. Generally, if a high prediction confidence score is obtained from the current model, then the image is considered correctly classified and the inference will be terminated immediately.

\begin{figure}
    \centering
    \includegraphics[width=0.47\textwidth]{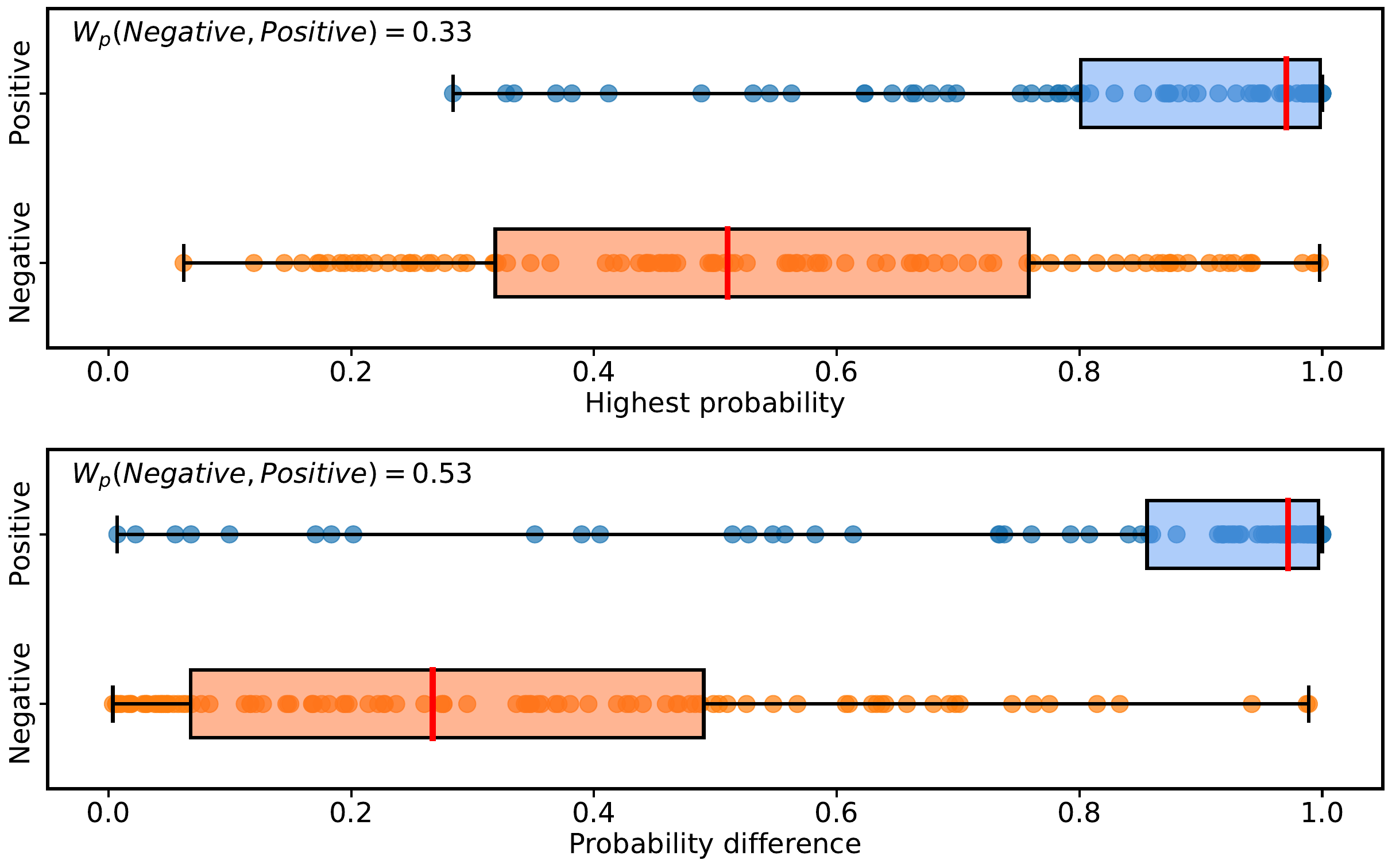}
    \caption{The distributions of negative results and positive results along different confidence metrics. $W_p$ denotes the Wasserstein distance between negative results and positive results. The larger the value of $W_p$, the more distinct the two distributions, such that our dynamic inference framework can more accurately determine whether the sample is correctly classified.}
    \label{fig:metrics}
\end{figure}

In this paper, we rethink the efficacy of the confidence metric. First, we randomly sample 100 negative prediction results and 100 positive prediction results from a well-trained model. Subsequently, we summarize the distribution of the sampled data along the highest prediction probability in Fig. \ref{fig:metrics} and exploit Wasserstein distance to quantify the similarity between the distributions of positive samples and negative samples. The smaller the Wasserstein distance between two distributions, the 
more similar the two distributions are. As shown in the upper figure of Fig. \ref{fig:metrics}, the negative samples and positive samples are distributed close along the highest probability with a small Wasserstein distance, which reveals that this confidence metric fails to effectively separate the positive predictions and negative predictions of a model, degrading the efficacy of dynamic inference. To address the issue, we introduce a novel metric, the probability difference, to control the termination of dynamic inference. The probability difference is defined as the difference between the highest prediction probability and the second highest probability, which is formulated as:

\begin{table}[tbp]
  \centering
  \caption{Comparison of different confidence metrics in terms of the trade-off between model complexity and accuracy. The accuracy is measured on ImageNet-1K.}
    \resizebox{0.47\textwidth}{!}{
        \begin{tabular}{lcrr}
        \toprule
        \toprule
        \textbf{Architecture} & \textbf{Metric}  & \textbf{\#MACs (B)} & \textbf{Top-1 Acc. (\%)} \\
        \midrule
        \multirow{4}[4]{*}{ResNet-50} & Highest probability    & 2.07  & 76.44 \\
             & Probability difference    & 2.07  & \textbf{76.68} \\
    \cmidrule{2-4}    & Highest probability    & 2.39  & 76.91 \\
            & Probability difference   & 2.33  & \textbf{77.07} \\
        \bottomrule
        \bottomrule
        \end{tabular}%
    }
  \label{tab:metrics}%
\end{table}%

\begin{equation}
\small
    \label{eq:difference}
    \mathcal{D}_{\mathcal{N}} = \mathcal{I}_{\mathcal{N}} - \mathcal{I}_{\mathcal{N}}'
\end{equation}
where $\mathcal{I}_{\mathcal{N}}'$ represents the second highest prediction probability. Equation \ref{eq:difference} reveals that, unlike existing confidence metric which only focuses on the highest prediction probability, the proposed confidence metric considers both the highest prediction itself and its advantages over other competitors. The distributions of the negative samples and positive samples along the probability difference are demonstrated in the lower figure of Fig. \ref{fig:metrics}, where we observe that the two distributions are more distinct and the Wasserstein distance between them is much larger compared to existing confidence metric (i.e., the highest probability), which indicates that the proposed confidence metric is able to estimate the correctness of a prediction more accurately, enabling effective control over the dynamic inference. 
\major{After evaluating the confidence of a prediction with the proposed metric, we will compare the evaluation result with a preset threshold value $\mathcal{D}_0$. If the evaluation result is larger than the preset threshold value, the prediction is considered confident and the inference will be terminated. Otherwise, the image will be sent to a larger model for more accurate prediction. By changing the threshold value $\mathcal{D}_0$, we are able to flexibly adjust the trade-off of the dynamic inference between inference overhead and accuracy. Specifically, a higher threshold value will force more images to flow to large models, and thus the accuracy will be improved at the cost of higher inference costs. On the contrary, reducing the threshold value will allow more images to exit at small models, thereby saving the inference overhead.
To validate the proposed confidence metric, we perform experiments on ImageNet-1K and present the results in TABLE \ref{tab:metrics}, where we observe that the proposed metric remarkably improves accuracy without sacrificing the computational cost.}

\subsubsection{Prediction Accumulation}
\label{subsec:accumulate}
During dynamic inference, hard samples may flow through multiple models. Some approaches directly adopt the output of the last model as the final result \cite{teerapittayanon2016branchynet}, which wastes the information from the previously executed models and consequently losses the opportunity to further improve accuracy. Instead, some other methods propose to utilize the information of previous models by merging the feature maps from previous models into the current model for higher accuracy \cite{yang2020resolution}. However, the fusion of feature maps of different models introduces additional computational overhead, reducing the efficiency of dynamic inference. 

\begin{figure}
    \centering
    \includegraphics[width=0.47\textwidth]{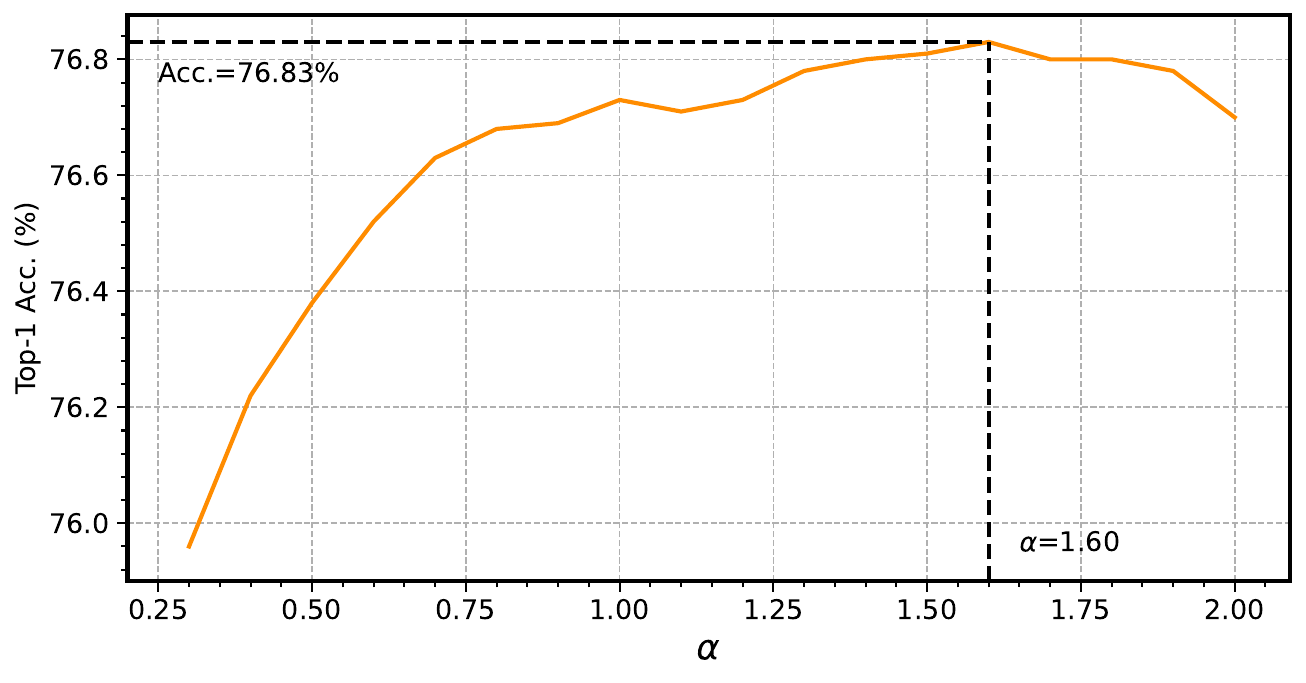}
    \caption{The impact of the value of $\alpha$ on the final accuracy of dynamic inference. We observe the highest accuracy at $\alpha=1.60$, and thus we fix $\alpha=1.60$ for subsequent experiments. The target dataset is ImageNet-1K.}
    \label{fig:alpha}
\end{figure}

\major{To address the above concerns, we propose prediction accumulation to effectively utilize the information from different models for higher accuracy. Different from the fusion of feature maps \cite{yang2020resolution} which requires a large amount of additional computation, we efficiently integrate the information from different models without compromising the computational overhead by accumulating the output of the last fully connected layer in each model (i.e., the logits), which is formulated as:}

\begin{equation}
\small
    \label{eq:accumulate}
    \mathcal{Z}_i' = \alpha \mathcal{Z}_i + \mathcal{Z}_{i-1}'
\end{equation}
where $\mathcal{Z}_i$ denotes the logits of the current model, and $\mathcal{Z}_i'$ represents the accumulated logits of the current model, which will be used to calculate the prediction of the current model. $\mathcal{Z}_{i-1}'$ is the accumulated logits of the previous model and $\alpha$ is a hyperparameter to control the contribution of the prediction of the current model (i.e., $\mathcal{Z}_i$). The value of $\alpha$ can affect the final accuracy of dynamic inference obviously. To identify the optimal value of $\alpha$, we sample multiple values of $\alpha$ and summarize the relationship between accuracy and $\alpha$ in Fig. \ref{fig:alpha}, where we observe the highest accuracy when $\alpha=1.60$. Therefore, we fix $\alpha=1.60$ in our experiments.

\begin{table}
  \centering
  \caption{The impact of our prediction accumulation strategy on model computation, inference latency, and accuracy. The inference latency is measured on Jetson Xavier, and the accuracy is measured on the ImageNet-1K dataset.}
    \resizebox{0.47\textwidth}{!}{
    \begin{tabular}{lcrrr}
    \toprule
    \toprule
    \textbf{Architecture} & \textbf{Accumulation} & \textbf{\#MACs (B)} & \textbf{Latency (ms)} & \textbf{Top-1 Acc. (\%)} \\
    \midrule
    \multirow{4}[4]{*}{ResNet-50} & \XSolidBrush    & 2.07  & 7.84  & 76.68 \\
          & \Checkmark   & 2.07  & 7.91  & \textbf{76.83} \\
\cmidrule{2-5}          & \XSolidBrush    & 2.33  & 8.63  & 77.07 \\
          & \Checkmark   & 2.33  & 8.60  & \textbf{77.35} \\
    \bottomrule
    \bottomrule
    \end{tabular}%
    }
  \label{tab:reuse}%
\end{table}%


\major{The logits of a model can directly determine the prediction results, and thus accumulating logits can effectively utilize the information from multiple models for higher accuracy. The overhead of accumulating logits is determined by the number of logits in each model and the number of models to accumulate. Specifically, the computational cost of logits accumulation can be calculated as follows: 
\begin{equation}
    \label{eq:acc_cost}
    Q_{acc} = n_l \cdot (n_m - 1)
\end{equation}
where $n_l$ is the number of logits and $n_m$ is the number of models. For example, for two models with 1,000 logits, the computational cost of logits accumulation will be $1,000 \times (2 - 1) = 1,000$ FLOPs, which can be neglected compared to the inference overhead of CNN backbones (e.g., ResNet-50 with 4.1 Billion FLOPs). As shown in the experimental results in TABLE \ref{tab:reuse}, the accumulation strategy improves the accuracy remarkably while not increasing the computational cost and inference latency of our framework.
}


\subsubsection{Dynamic Inference Algorithm}
\label{subsubsec:algorithm}

We demonstrate our dynamic inference algorithm in detail in Algorithm \ref{alg:one}. Given a series of CNN models ordered from low to high computational complexity and an input image, we initiate the dynamic inference with the smallest model and gradually activate models with higher computational complexity. For each model, we first perform inference with the model to obtain its prediction logits, then, we accumulate the logits of this model with all previously executed models as Equation \ref{eq:accumulate}. Subsequently, the proposed termination metric of the current model is calculated using the accumulated logits according to Equation \ref{eq:difference}, which is then compared with the preset threshold $\mathcal{D}_0$. 
\minor{If the calculated metric is larger than the threshold, the predicted result is considered confident and dynamic inference will be terminated immediately. Otherwise, a larger model will be activated for inference. In practice, we observe that, for most images, dynamic inference is able to produce a confident prediction and be terminated at the smallest model. In this case, the overall latency of dynamic inference is equal to the inference latency of the smallest model. Consequently, we avoid using large models for most images, saving computation and reducing latency significantly compared to static inference. The results are presented in the Experimental Results section.}

\begin{algorithm}[t]
\footnotesize
\caption{\small Dynamic Inference Algorithm}\label{alg:one}
\KwData{CNN models \{$\mathcal{N}_1$, $\mathcal{N}_2$, ..., $\mathcal{N}_n$\}, input image $x$, termination threshold $D_0$ 
}
\KwResult{Prediction result $o$}
$i \gets 0$\;
\While{$i < n$}{
    $\mathcal{Z}_i \gets \mathcal{N}_i(x)$ \ \Comment{\scriptsize Inference with the current model}
    \uIf{$i=0$}{
        $\mathcal{Z}_i' \gets \mathcal{Z}_i$
    }
    \Else{
        $\mathcal{Z}_i' \gets AccumulateLogits(\mathcal{Z}_i,\ \mathcal{Z}_{i-1}')$ \ \Comment{\scriptsize See Eq. \ref{eq:accumulate}}
    }
    \Comment{\scriptsize Calculate the termination metric, see Eq. \ref{eq:difference}}
    $\mathcal{D}_{\mathcal{N}_i} \gets CalculateMetric(\mathcal{N}_i,\ \mathcal{Z}_i')$ \\
    \If{$\mathcal{D}_{\mathcal{N}_i} > \mathcal{D}_0 $ }{
    Break \ \Comment{Terminate dynamic inference}
    }
    $i \gets i+1$ \Comment{Activate the next model}
}
$o \gets Softmax(\mathcal{Z}_i')$ \ \Comment{Calculate the final result}
\end{algorithm}

\section{Experimental Results}
\label{sec:exp}

In this section, we perform extensive experiments on different benchmarks to validate the efficacy of \edge and demonstrate its advantages over existing SOTA approaches in terms of accuracy, computational complexity (i.e., MACs), and run-time efficiency. Further, we conduct ablation study to show the contribution of each component in our framework.

\begin{table}[hbp]
  \centering
  \caption{Hardware specifications of three platforms. The column ``\#Cores" denotes the number of CUDA cores and CPU cores for GPU platforms (i.e., AGX Xavier and Jetson Nano) and the CPU platform (I7-9750H), respectively.}
    \resizebox{0.47\textwidth}{!}{
    \begin{tabular}{lrrrrr}
    \toprule
    \toprule
    \textbf{Device} & \textbf{Power} & \textbf{Memory} & \textbf{\#Cores} & \textbf{Core Freq.} & \textbf{Performance}  \\
    \midrule
    AGX Xavier & 15 W & 32 GB & 512 & 900 MHz & 11.0 TOPS  \\
    \midrule
    Jetson Nano  & 5 W  & 4 GB & 128 & 992 MHz & 0.5 TOPS  \\
    \midrule
    i7-9750H & 45 W & 16 GB & 6 & 2600 MHz & 0.4 TOPS \\
    \bottomrule
    \bottomrule
    \end{tabular}%
    }
  \label{tab:device}%
\end{table}%

\subsection{Hardware Devices}
\label{subsec:hardware}
To validate the run-time efficiency of \edge, including inference latency and throughput, we select two representative embedded GPU platforms, NVIDIA AGX Xavier and Jeton Nano, and Intel i7-9750H@2.6GHz CPU to deploy different methods and compare their performance. 
The specifications of selected devices are shown in TABLE \ref{tab:device}.


\subsection{Datasets}
\label{subsec:dataset}
\major{We validate the proposed \edge on four representative datasets: 1) CIFAR-10, 2) CIFAR-100, 3) ImageNet-100, and 4) ImageNet-1K \cite{deng2009imagenet}. ImageNet-1K (also known as ImageNet or ILSVRC-2012) is one of the most popular large-scale datasets for image classification, which includes 1,000 classes.} ImageNet-100 is a subset of ImageNet-1K, which consists of 100 classes randomly selected from ImageNet-1K. The details of ImageNet-100 can be found in the code repository. All images are preprocessed following a simple configuration as \cite{radosavovic2020designing}.

\subsection{Networks}
\label{subsec:network}
\major{We apply our \edge framework to three widely utilized CNN backbones, VGG16\_BN \cite{simonyan2014very}, ResNet-50 \cite{he2016deep}, and RegNet-X \cite{radosavovic2020designing}.} For each model, we employ \edge to remove the spatial redundancy in input images and the architecture redundancy in networks, thereby reducing the computational cost and improving the inference efficiency. As a comparison, we also report the results of other methods.

\begin{figure}[tbp]
    \centering
    \includegraphics[width=0.48\textwidth]{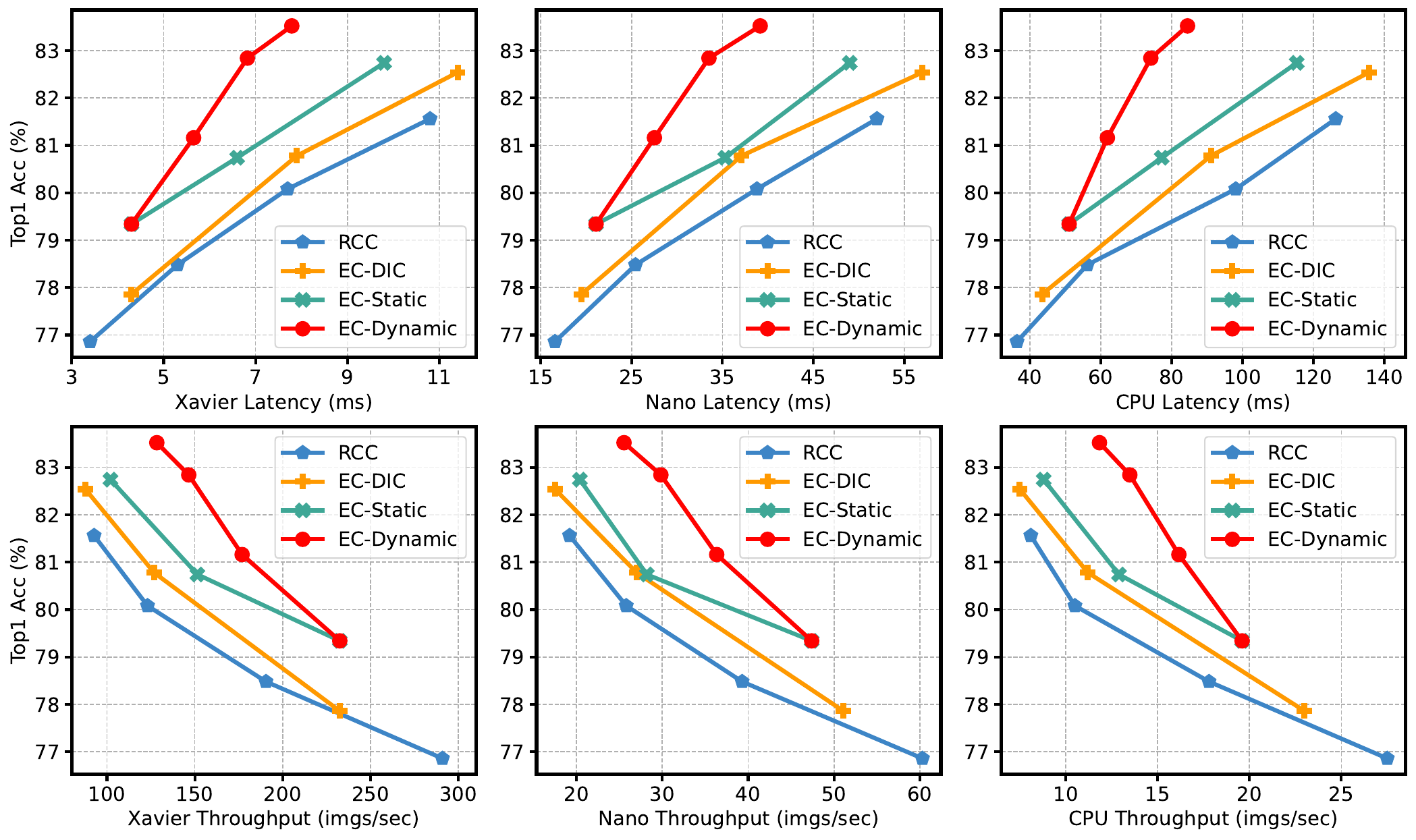}
    \caption{The real performance of ResNet-50 compressed by different methods on three distinct hardware devices. Accuracy is measured on ImageNet-100.}
    \label{fig:res50_perf}
\end{figure}

\subsection{Optimization Settings}
\label{subsec:setting}

All models in our experiments are trained using SGD optimizer with a momentum of 0.9. We first train models for 100 epochs without using dynamic image cropping, where the first 5 epochs are for warmup. \major{For experiments on ImageNet-100 and ImageNet-1K,} the learning rate is set to 2.0, which will be decayed by exponential learning rate policy with a decay factor of 0.02. The training batch size is set to 1024. Subsequently, the proposed dynamic image cropping is utilized to fine-tune the pretrained models for 20 epochs. The learning rate for fine-tuning is 5e-4. In addition, we also use label smoothing with the smoothing factor $\epsilon=0.1$ \cite{szegedy2016rethinking} to prevent overfitting. \major{For experiments on CIFAR-10 and CIFAR-100, the initial learning rate is 0.1 and the training batch size is 128.}

\begin{table}[tbp]
  \centering
  \caption{Results of ResNet-50 on ImageNet-100. RCC-Baseline represents the baseline ResNet-50 model, where we crop and resize all images to the size 224$\times$224 with RCC.}
    \resizebox{0.47\textwidth}{!}{
    \begin{tabular}{lrrrr}
    \toprule
    \toprule
    \textbf{Method} & \textbf{\#Params (M)} & \textbf{\#MACs (B)} & \textbf{$\downarrow$ MACs (\%)}& \textbf{Top1 Acc. (\%)} \\
    \midrule
    \rcc-Baseline & 23.7  & 4.1 & 0.0  & 81.6  \\
    \ssd    & 24.0  & 4.2 & -2.4  & 82.5 \\
    \css & 17.3 & 3.0 & 26.8 & 82.7 \\
    \textbf{EC-Dynamic} & \textbf{13.6} & \textbf{2.0} & \textbf{51.2} & \textbf{83.5} \\
    \midrule
    \rcc   & 23.7  & 3.0  & 26.8 & 80.1  \\
    \ssd    & 24.0  & 2.6  & 36.6  & 80.8  \\
    \css & 14.5 & 2.4 & 41.5 & 81.5 \\
    \textbf{EC-Dynamic} & \textbf{11.8} & \textbf{1.7} & \textbf{58.5} & \textbf{82.8} \\
    \midrule
    \rcc   & 23.7  & 1.1 & 73.2   & 76.9  \\
    \ssd   & 24.0  & 1.2 & 70.7  & 77.9 \\
    \css & \textbf{7.8} & \textbf{1.0} & \textbf{75.6} & 79.3 \\
    \textbf{EC-Dynamic} & 8.9 & 1.2 & 70.7 & \textbf{80.2} \\
    \bottomrule
    \bottomrule
    \end{tabular}%
    }
  \label{tab:res-100}%
\end{table}%

\subsection{Evaluation Methodology}
\label{subsec:eval_methodology}
In this paper, we propose three novel approaches to comprehensively reduce the computational cost of CNNs. Thanks to the flexible design of these approaches, they can be used separately or coupled for a higher compression ratio. To better demonstrate the flexibility and efficacy of our design, we evaluate different combinations of the three approaches. Specifically, \ssd represents that only the dynamic image cropping component is exploited, while \css denotes both the dynamic image cropping and compound shrinking are adopted. Finally, EC-Dynamic denotes the completed framework that contains all three components, which further integrates the dynamic inference approach based on EC-Static.

\subsection{Results on ImageNet-100}
We conduct experiments on ImageNet-100 with ResNet-50 and RegNet-X, where we use different methods mentioned in Subsection \ref{subsec:eval_methodology} to compress models to different complexities. As a comparison, we use the most popular image cropping method, ResizedCenterCrop (\rcc) to crop and resize images to different sizes. Finally, all models are deployed to selected hardware to evaluate their latency and throughput.

\subsubsection{ResNet-50}
As shown in TABLE \ref{tab:res-100}, all of our approaches outperform the competitor (i.e., RCC) in terms of the model complexity, on-device execution efficiency, and accuracy. Specifically, compared to the baseline ResNet-50 (\rcc-Baseline), \ssd improves the accuracy by 0.9\% with a negligible increase in model parameters (1.2\%) and MACs (2.4\%), while \css further pushes up the accuracy improvement to 1.1\% with a parameter reduction of 27.0\% and a MACs reduction of 26.8\%. Finally, EC-Dynamic achieves the best performance, which compresses the MACs by 51.2\% while still improving the accuracy by 1.9\% compared to RCC-Baseline.
In the low complexity regime, EC-Dynamic achieves 3.3\% higher accuracy than \rcc with only 37.6\% model parameters (8.9M v.s. 23.7M) and similar MACs. Meanwhile, Fig. \ref{fig:res50_perf} indicates that all of our methods outperform \rcc by a large margin across a wide spectrum of inference latency and throughput on different resource-constrained embedded devices. Particularly, EC-Dynamic achieves 83.5\% top-1 accuracy with a latency of 7.8 ms on Xavier, which is 1.9\% higher in accuracy and 27.7\% lower in latency compared to \rcc (81.6\% top1 accuracy, 10.8 ms). At the same time, the throughput of EC-Dynamic on Xavier is 128.4 \textit{imgs/sec}, which is 38.5\% higher than \rcc (92.7 \textit{imgs/sec}). On Nano and Intel i7-9750H CPU, EC-Dynamic also improves the throughput by 33.0\% and 46.2\%, and reduces the latency by 24.7\% and 33.1\%, respectively.

\begin{figure}[tbp]
    \centering
    \includegraphics[width=0.48\textwidth]{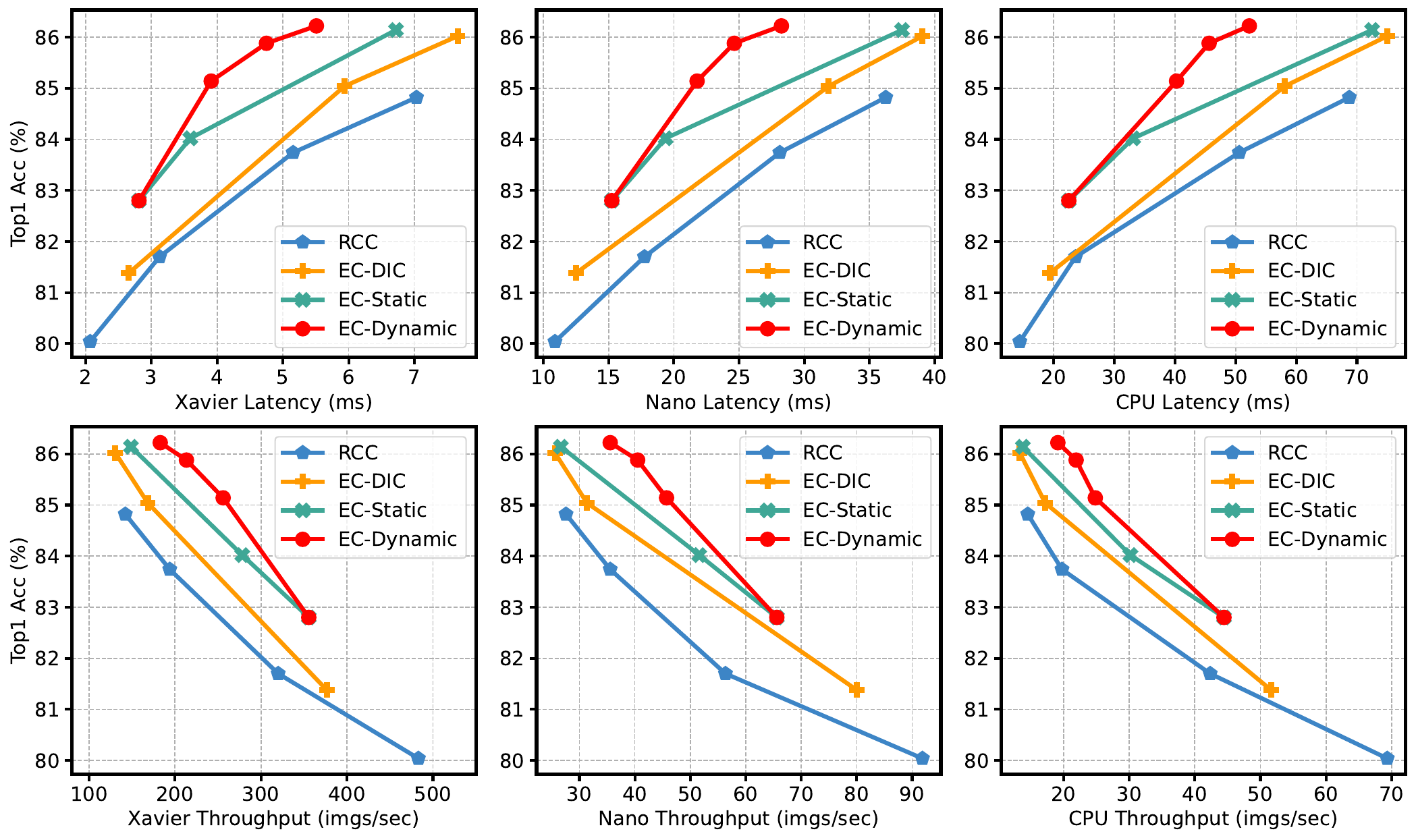}
    \caption{The real performance of RegNet-X compressed by different methods on three distinct hardware devices. Accuracy is measured on ImageNet-100.}
    \label{fig:reg_perf}
\end{figure}

\begin{table}
  \centering
  \caption{Results of RegNet-X on ImageNet-100. RCC-Baseline represents the baseline RegNet-X model with all input images cropped and resized to 224$\times$224 with RCC.}
    \resizebox{0.47\textwidth}{!}{
    \begin{tabular}{lrrrr}
    \toprule
    \toprule
    \textbf{Method} & \textbf{\#Params (M)} & \textbf{\#MACs (B)} & \textbf{$\downarrow$ MACs (\%)} & \textbf{Top1 Acc. (\%)} \\
    \midrule
    \rcc-Baseline & 8.4 & 1.6 & 0.0 & 84.8 \\
    \ssd & 8.7 & 1.3 & 18.8 & 85.0 \\
    \css & 6.3 & 1.3 & 18.8 & 86.1 \\
    \textbf{EC-Dynamic} & \textbf{4.4} & \textbf{0.8} & \textbf{50.0} & \textbf{86.2} \\
    \midrule
    \rcc   & 8.4 & 0.7 & 56.3 & 82.3 \\
    \ssd & 8.7 & 0.8 & 50.0 & 83.8 \\
    \css & 3.6 & 0.6 & 62.5 & 84.0 \\
    \textbf{EC-Dynamic} & \textbf{3.3} & \textbf{0.6} & \textbf{62.5} & \textbf{85.1} \\
    \midrule
    \rcc   & 8.4 & 0.4 & 75.0 & 80.0 \\
    \ssd & 8.7 & 0.5 & 68.8 & 81.4 \\
    \css & \textbf{2.5} & \textbf{0.4} & \textbf{75.0} & 82.8 \\
    \textbf{EC-Dynamic} & 2.8 & 0.5 & 68.8 & \textbf{83.7} \\
    \bottomrule
    \bottomrule
    \end{tabular}%
    }
  \label{tab:regnet}%
\end{table}%

\subsubsection{RegNet-X}
The experimental results of RegNet-X are summarized in TABLE \ref{tab:regnet} and Fig. \ref{fig:reg_perf}, where we also observe a significant improvement of our method. EC-Dynamic outperforms the baseline RegNet-X (RCC-Baseline) with an improvement of 1.4\% in accuracy and a reduction of 50.0\% in MACs. Meanwhile, EC-Dynamic reduces the model parameters by 47.6\% (4.4M v.s. 8.4M). In the low MACs regime, EC-Dynamic observes a remarkable 3.7\% improvement in accuracy with only 33.3\% parameters (2.8M v.s. 8.4M) compared to \rcc. As for the real performance on hardware, EC-Dynamic obtains an accuracy of 86.2\% with 5.5 ms latency on Xavier, which is 1.4\% higher in accuracy and 21.4\% lower in latency than \rcc (84.8\% top-1 accuracy, 7.0 ms). Similarly, the latency reductions of EC-Dynamic on Nano and Intel CPU are 22.0\% and 24.0\%, respectively. Besides, \css also observes a 29.0\% throughput improvement (35.6 \textit{imgs/sec} v.s. 27.6 \textit{imgs/sec}) on Nano and a 31.7\% throughput improvement (19.1 \textit{imgs/sec} v.s. 14.5 \textit{imgs/sec}) on CPU compared to \rcc.

\begin{figure}
    \centering
    \includegraphics[width=0.48\textwidth]{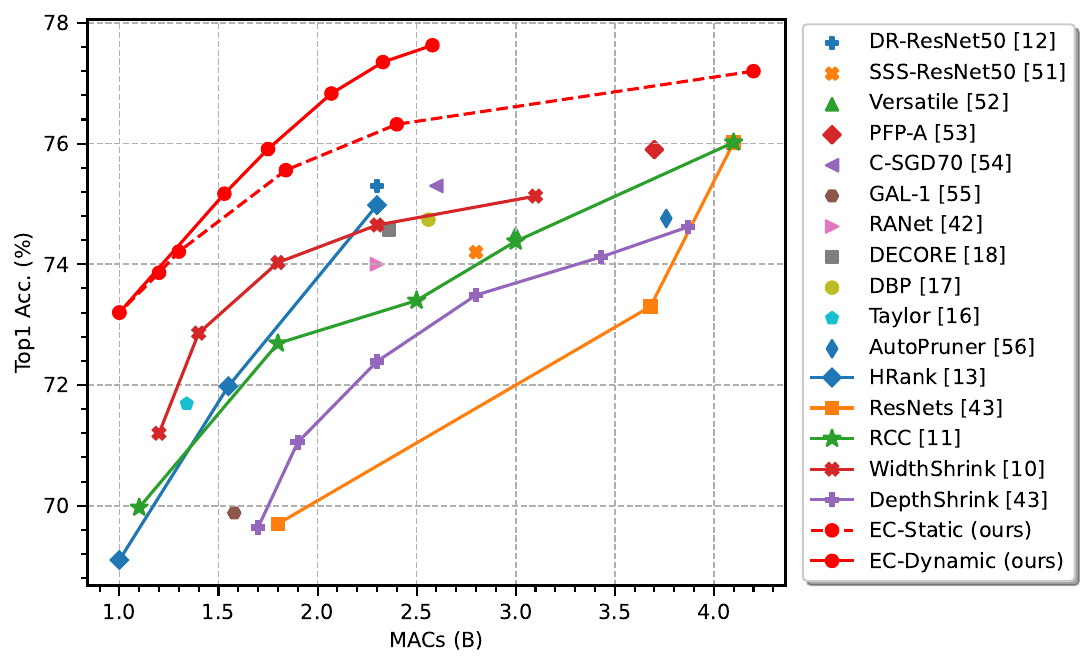}
    \caption{Comparison of our \edge with other state-of-the-art model compression methods. The baseline model is ResNet-50 and the dataset is ImageNet-1K.}
    \label{fig:res-1k}
\end{figure}

\subsection{Results on ImageNet-1K}
In this subsection, we evaluate our approach on ImageNet-1K, and compare the evaluation results with many SOTA CNN compression frameworks. To enable a comprehensive comparison with more SOTA frameworks, we employ ResNet-50 as the baseline network. In addition, we also compare our compressed models with many popular backbone architectures in different computation regimes.

\subsubsection{Comparison with \sota Compression Methods}
Starting with the baseline ResNet-50 model, we implement different model shrinking methods, including resolution shrinking (\rcc), width shrinking (WidthShrink) \cite{sandler2018mobilenetv2, zagoruyko2016wide}, depth shrinking (DepthShrink) \cite{he2016deep}, \ssd, \css, and EC-Dynamic to compress the three dimensions of the model to different MACs regimes and compare their performance. In addition, we also report the performance of multiple \sota model compression techniques from the related papers, including DR-ResNet50 \cite{zhu2021dynamic}, SSS-ResNet50 \cite{huang2018data}, Versatile \cite{wang2018learning}, PFP-A \cite{liebenwein2019provable}, C-SGD70 \cite{ding2019centripetal}, GAL-1 \cite{lin2019towards}, HRank \cite{lin2020hrank}, AutoPruner \cite{luo2020autopruner}, and RANet \cite{yang2020resolution}. The comparison results are summarized in Fig. \ref{fig:res-1k}, which shows that our method achieves the highest accuracy across a wide range of MACs. Particularly, compared to the baseline ResNet-50, our \css achieves 1.2\% accuracy improvement (76.0\% to 77.2\%) with a negligible increase in MACs (4.1B to 4.2B). Moreover, EC-Dynamic further improves the accuracy to 77.6\% with only 2.6B MACs, which is 1.6\% higher in accuracy and 36.6\% lower in MACs compared to the baseline ResNet-50. As we continue to reduce the MACs budget, EC-Dynamic reduces the MACs by 48.8\% (4.1B to 2.1B) while still achieving 0.8\% higher accuracy (76.0\% to 76.8\%). In the lowest MACs regime, EC-Dynamic and EC-Static achieve similar trade-offs between model MACs and accuracy, both of which remarkably improve the accuracy by 4.2\% (70.0\% to 74.2\%) compared to \rcc with similar MACs. In comparison with other \sota compression methods, our method also achieves the best trade-off between MACs and accuracy. For example, EC-Dynamic achieves 5.3\% higher accuracy (75.2\% v.s. 69.9\%) than GAL-1 \cite{lin2019towards} with less MACs (1.5B v.s. 1.6B).

\begin{table}[tbp]
  \centering
  \caption{Comparison with other popular backbone networks.}
    \resizebox{0.47\textwidth}{!}{
        \begin{tabular}{lrrrr}
        \toprule
        \toprule
        \textbf{Model} & \textbf{\#Params (M)} & \textbf{\#MACs (B)} & \textbf{$\downarrow$ MACs (\%)} & \textbf{Top1 Acc. (\%)} \\
        \midrule
        ResNet-50\cite{he2016deep} & 25.6 & 4.1 &   0.0   & 76.0 \\
        \midrule
        ResNet-101\cite{he2016deep} & 44.6 & 7.9 &    -92.7   & 77.4 \\
        DenseNet-161\cite{huang2017densely} & 28.7 & 7.9 &   -92.7    & 77.1 \\
        InceptionV3\cite{szegedy2016rethinking} & 27.2  & 5.8 &   -41.5   & 77.3 \\
        \ssd & 25.9 & 4.2 &  -2.4    & 77.2 \\
        \textbf{EC-Dynamic} & \textbf{20.4} & \textbf{2.6} & \textbf{36.6} & \textbf{77.6} \\
        \midrule
        ResNet34\cite{he2016deep} & 21.8  & 3.7  &    9.8   & 73.3 \\
        DenseNet-169\cite{huang2017densely} & \textbf{14.2} & 3.4  &   17.1    & 75.6 \\
        \ssd & 25.9  & 3.1  &   24.4      & 76.3 \\
        \css &  15.4      &  2.4     &    41.5   & 76.3  \\
        \textbf{EC-Dynamic} & 17.1 & \textbf{2.1}  & \textbf{48.8} & \textbf{76.8} \\ 
        \midrule
        ResNet-18\cite{he2016deep} & 11.7  & 1.8  &   56.1    & 69.8 \\
        DenseNet-121\cite{huang2017densely} & \textbf{8.0} & 2.9  &   29.3    & 74.6 \\
        BN-Inception\cite{ioffe2015batch} & 11.2  & 2.1  &   48.8     & 73.5 \\
        \ssd & 25.9  & 1.9  &     53.7  & 74.9 \\
        \css & 13.5    &  1.8     &  56.1    & 75.6  \\
        \textbf{EC-Dynamic} & 15.1 & \textbf{1.8} & \textbf{ 56.1} & \textbf{75.9} \\
        \bottomrule
        \bottomrule
        \end{tabular}%
    }
  \label{tab:backbones}%
\end{table}%

\begin{table}[htbp]
  \centering
  \caption{Comparison with other dynamic inference frameworks. \{d, w, r\} denote the dimensions involved for dynamic inference.}
    \resizebox{0.47\textwidth}{!}{
    \begin{tabular}{lcccrr}
    \toprule
    \toprule
    \textbf{Method} & \textbf{d} & \textbf{w} & \textbf{r} & \textbf{\#MACs (B)} & \textbf{Top-1 Acc. (\%)} \\
    \midrule
    ResNet-50 \cite{he2016deep} &       &       &       & 4.1  & 76.0 \\
    \midrule
    SkipNet \cite{wang2018skipnet} &\Checkmark    &       &       & 3.6  & 76.2 \\
    ConvNet-AIG \cite{veit2018convolutional} &\Checkmark    &       &       & 3.1  & 76.2 \\
    Channel Selection \cite{herrmann2020channel} &       &\Checkmark    &       & \textbf{2.5}  & 76.2 \\
    DR-ResNet \cite{zhu2021dynamic} &       &       &\Checkmark    & 3.2  & 77.0 \\
    \textbf{EC-Dynamic (ours)} &\Checkmark    &\Checkmark    &\Checkmark    & 2.6  & \textbf{77.6} \\
    \midrule
    RANet \cite{yang2020resolution} &\Checkmark    &       &\Checkmark    & \textbf{2.0}  & 75.2 \\
    ConvNet-AIG \cite{veit2018convolutional} &\Checkmark    &       &       & 2.6  & 75.3 \\
    DR-ResNet \cite{zhu2021dynamic} &       &       &\Checkmark    & 2.3  & 75.3 \\
    MSDNet \cite{huang2018multiscale} &\Checkmark    &       &       & 2.1  & 75.7 \\
    Channel Selection \cite{herrmann2020channel} &       &\Checkmark    &       & 2.3  & 76.1 \\
    \textbf{EC-Dynamic (ours)} &\Checkmark    &\Checkmark    &\Checkmark    & 2.1  & \textbf{76.8} \\
    \bottomrule
    \bottomrule
    \end{tabular}%
    }
  \label{tab:exp_dynamic}%
\end{table}%

\subsubsection{Comparison with Popular Backbones}
In this experiment, we compare our results on ResNet-50 with other models from the ResNet family, such as ResNet-101 and ResNet-34, etc. In addition, we also conduct extensive comparisons with other popular backbones like DenseNets \cite{huang2017densely} and the Inception family \cite{szegedy2016rethinking, ioffe2015batch}. As demonstrated in TABLE \ref{tab:backbones}, in the highest MACs regime, EC-Dynamic uses 67.1\% less MACs (2.6B v.s. 7.9B) to achieve 0.5\% higher accuracy than DenseNet-161, while in the lowest MACs regime, EC-Dynamic also obtains the highest top-1 accuracy (75.9\%), which is 6.1\% and 2.4\% higher than ResNet-18 (69.8\%) and BN-Inception (73.5\%), respectively. The comparison results with other backbones reveal that our method can achieve promising results without redesigning the network architecture, which avoids the extremely time-consuming exploration of the architecture design space.

\subsubsection{Comparison with SOTA Dynamic Inference Frameworks}

We can observe from the above experiments that our compression framework with dynamic inference (i.e., EC-Dynamic) surpasses the one without dynamic inference (i.e., EC-Static) in model efficiency and accuracy. To further validate the advantages of our dynamic inference framework, we compare it with multiple SOTA dynamic inference frameworks. The comparison results are shown in TABLE \ref{tab:exp_dynamic}, from which we observe that our EC-Dynamic framework achieves significant improvements on accuracy without sacrificing model complexity compared to other approaches. It is worth noting that the most of existing dynamic inference approaches only adjust a single dimension during inference, while our approach enables the joint adaptation of the three dimensions based on our multi-dimensional compression framework, achieving higher accuracy and efficiency. This also reveals that all components of our framework can be seamlessly coupled for a better result.

\subsubsection{On-Device Efficiency of Dynamic Inference}
\minor{In this experiment, we demonstrate the running efficiency of our approach on various edge devices and analyze how the preset threshold affects the on-device latency and accuracy. As shown in TABLE \ref{tab:latency}, we first apply a small threshold (i.e., $0.1$) to our dynamic algorithm, which achieves higher accuracy and lower latency than static inference. As we increase the threshold, there are more images whose prediction confidence is smaller than the threshold, and thus more images are sent to the larger model for further inference. Consequently, both the classification accuracy and average inference latency of processing one image increase.}


\begin{table}[tbp]
  \centering
  \caption{Static inference v.s. Dynamic inference in terms of runtime latency and accuracy. The latency is represented by the average inference latency of all images.}
    \resizebox{0.47\textwidth}{!}{
    \begin{tabular}{lcrrrrr}
    \toprule
    \toprule
    \textbf{Method}  & \textbf{Threshold}  & \textbf{Xavier (ms)} & \textbf{Nano (ms)} & \textbf{CPU (ms)} & \textbf{Top1 (\%)} \\
    \midrule
    EC-Static & N.A.  & 8.3  & 41.6 & 81.8 & 75.6 \\
    EC-Dynamic &  0.1 & \textbf{7.0}  & \textbf{34.0} & \textbf{74.8} & \textbf{75.9} \\
    \midrule
    EC-Static & N.A.  & 9.4  & 49.4 & 96.9 & 76.3 \\
    EC-Dynamic  & 0.2 & \textbf{7.9}  & \textbf{38.2} & \textbf{84.9} & \textbf{76.8} \\
    \midrule
    EC-Static & N.A.   & 10.8  & 56.0 & 117.6 & 76.8 \\
    EC-Dynamic &  0.3 & \textbf{8.6}  & \textbf{41.6} & \textbf{93.0} & \textbf{77.4} \\
    \midrule
    EC-Static & N.A.  & 11.8 & 57.1 & 134.5 & 77.2 \\
    EC-Dynamic & 0.4 & \textbf{9.5}  & \textbf{44.7} & \textbf{100.4} & \textbf{77.6} \\
    \bottomrule
    \bottomrule
    \end{tabular}%
    }
  \label{tab:latency}%
\end{table}%

\subsection{\major{Results on CIFAR-10 and CIFAR-100}}

\major{To better demonstrate the efficacy of our approach on small-scale datasets, we conduct extensive experiments on both CIFAR-10 and CIFAR-100 datasets. The experimental results are shown in TABLE \ref{tab:cifar}, which indicate that our approach has significant advantages over other existing methods on both datasets. For instance, our approach observes 2.09\% higher top-1 accuracy with 41.13\% less computation on CIFAR-10.}


\begin{table}[htbp]
  \centering
  \caption{The experimental results on CIFAR-10 and CIFAR-100. The baseline network is VGG-16 with batch normalization layers. \textbf{CR} in the table denotes the \textbf{C}ompression \textbf{R}atio.}
    \resizebox{0.47\textwidth}{!}{
        \begin{tabular}{llrrr}
    \toprule
    \toprule
    \textbf{Dataset} & \multicolumn{1}{l}{\textbf{Method}} & \multicolumn{1}{c}{\textbf{\#MACs (M)}} & \multicolumn{1}{c}{\textbf{MACs CR (\%)}} & \multicolumn{1}{c}{\textbf{Top1 Acc. (\%)}} \\
    \midrule
    \multirow{8}[6]{*}{CIFAR-10} & VGG16\_BN \cite{simonyan2014very} & 313.74 & 0.00  & 93.96 \\
\cmidrule{2-5}          & GAL-0.1 \cite{lin2019towards} & 171.89 & 45.21 & 90.73 \\
          & Hrank \cite{lin2020hrank} & 108.61 & 65.38 & 92.34 \\
          & \textbf{EdgeCompress (ours)} & \textbf{101.18} & \textbf{67.75} & \textbf{92.82} \\
\cmidrule{2-5}          & SSS \cite{huang2018data}   & 183.13 & 41.63 & 93.02 \\
          & Zhao et al \cite{zhao2019variational} & 190.00 & 39.44 & 93.18 \\
          & Hrank \cite{lin2020hrank} & 145.61 & 53.59 & 93.43 \\
          & \textbf{EdgeCompress (ours)} & \textbf{120.55} & \textbf{61.58} & \textbf{93.64} \\
    \midrule
    \multirow{3}[2]{*}{CIFAR-100} & SSS \cite{huang2018data}   & 223.13 & 28.89 & 71.08 \\
          & Zhao et al \cite{zhao2019variational} & 256.00 & 18.42 & 73.33 \\
          & \textbf{EdgeCompress (ours)} & \textbf{206.34} & \textbf{34.24} & \textbf{73.35} \\
    \bottomrule
    \bottomrule
    \end{tabular}%
    }
  \label{tab:cifar}%
\end{table}%

\subsection{\major{Robustness Analysis}}

\major{To evaluate the robustness of the proposed framework, we perform experiments in two long-tail settings: 1) exponential and 2) step, and compare the results with the normal setting. The experiments are conducted on CIFAR-10 with an unbalancing factor of 0.5. The experimental results are shown in TABLE \ref{tab:robust}, where the minor accuracy degradation in long-tail settings validates the robustness of our framework.}

\begin{table}[tbp]
  \centering
  \caption{Results in different long-tail settings, where "Normal" denotes the results in the normal setting. The network utilized is VGG16\_BN and the dataset is CIFAR-10.}
    \resizebox{0.47\textwidth}{!}{
        \begin{tabular}{lrrr}
        \toprule
        \toprule
        \textbf{Long-tail settings} & \multicolumn{1}{c}{\textbf{\#Params (M)}} & \multicolumn{1}{c}{\textbf{\#MACs (M)}} & \multicolumn{1}{c}{\textbf{Top1 Acc. (\%)}} \\
        \midrule
        VGG16\_BN \cite{simonyan2014very}  & 14.98 & 313.74 & 93.96 \\
        \midrule
        Normal & 5.39  & 101.18 & 92.82 \\
        Exponential & 5.92  & 112.12 & 91.69 \\
        Step  & 5.79  & 109.33 & 91.86 \\
        \midrule
        Normal & 6.59  & 126.14 & 93.54 \\
        Exponential & 6.90  & 132.54 & 92.42 \\
        Step  & 6.71  & 128.53 & 92.43 \\
    \bottomrule
    \bottomrule
    \end{tabular}%
    }
  \label{tab:robust}%
\end{table}%

\subsection{Analytical Experiments}
\label{subse:analytical}

In this subsection, we show the impact of some important hyperparameters on the final performance of our framework.

\subsubsection{The Architecture of The Foreground Predictor}

\major{The design of the foreground predictor can significantly affect the efficiency and accuracy of our framework. To validate the proposed lightweight foreground predictor, we conduct comparison experiments by integrating different advanced CNN into our framework as the foreground predictor. The experimental results are summarized in TABLE \ref{tab:predictors}, where we observe that our design achieves the best trade-off between accuracy and efficiency. Even though some modern architectures can achieve slightly higher accuracy, they result in magnitudes higher model complexity and latency. For instance, EfficientNet-B1 only achieves a mere 0.05\% accuracy improvement with 24.6$\times$ parameters and 6.3$\times$ MACs compared to our predictor, which significantly reduces the efficiency of the whole framework.}

\begin{table}[tbp]
  \centering
  \caption{\major{Comparison with different foreground predictors in terms of the classification accuracy and model efficiency.}}
    \resizebox{0.47\textwidth}{!}{
    \begin{tabular}{lrrrr}
    \toprule
    \toprule
    \textbf{Model} & \textbf{\#Params (M)} & \textbf{\#MACs (B)} & \textbf{Latency (ms)} & \textbf{Top1 Acc. (\%)} \\
    \midrule
    ResNet-18 \cite{he2016deep} & 11.23 & 1.81  & 3.01 & 75.59 \\
    ResNet-34 \cite{he2016deep} & 21.33 & 3.66  & 5.19 & 75.50 \\
    RegNet-X\_800MF \cite{radosavovic2020designing} & 6.65  & 0.80  & 4.50 & 75.50 \\
    RegNet-X\_1.6GF \cite{radosavovic2020designing} & 8.37  & 1.60  & 7.26 & 75.61 \\
    EfficientNet-B0 \cite{tan2019efficientnet} & 4.13  & 0.38  & 4.31 & 75.57 \\
    EfficientNet-B1 \cite{tan2019efficientnet} & 6.64  & 0.57  & 6.21 & \textbf{75.62} \\
    \textbf{Ours} & \textbf{0.27} & \textbf{0.09} & \textbf{1.04} & 75.57 \\
    \bottomrule
    \bottomrule
    \end{tabular}%
    }
  \label{tab:predictors}%
\end{table}%

\subsubsection{The Number of Models}
\begin{figure}[tbp]
    \centering
    \includegraphics[width=0.47\textwidth]{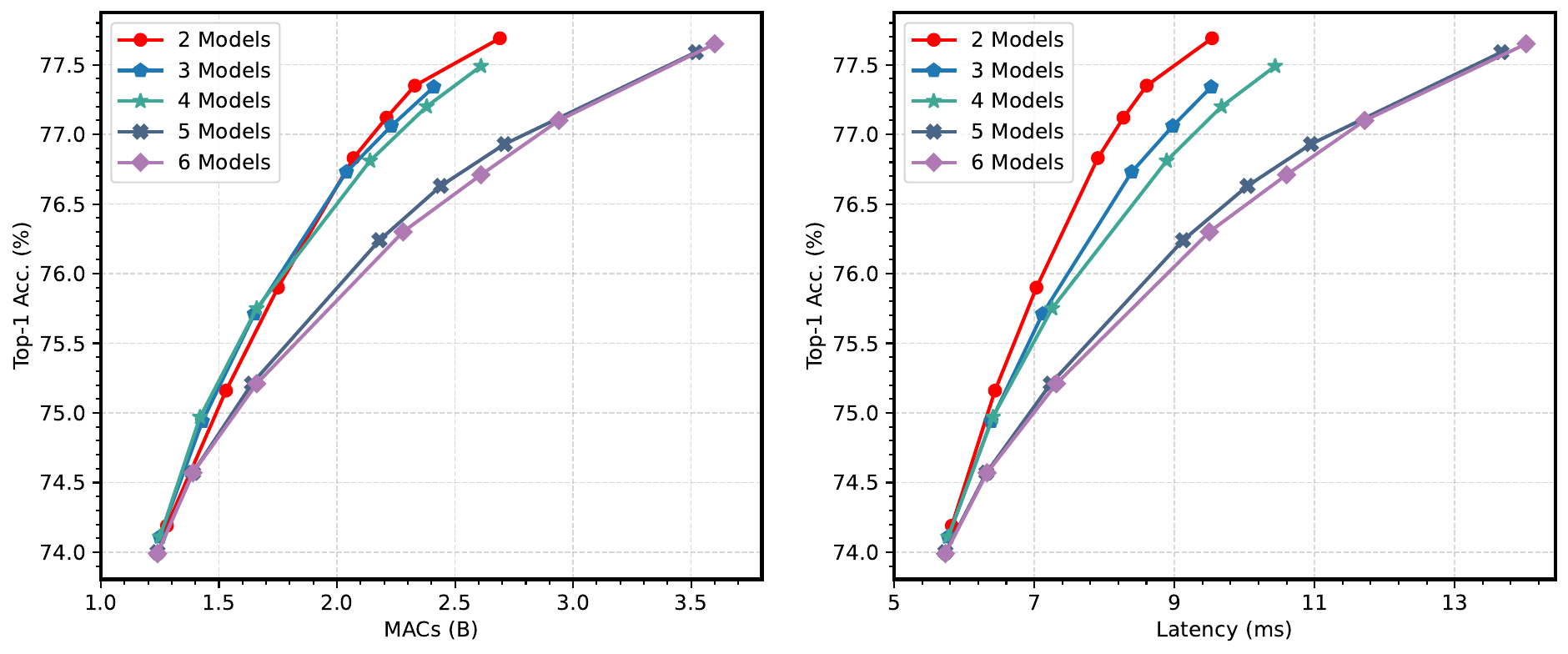}
    \caption{The impact of the number of models used for dynamic inference on the computational complexity and on-device latency. The latency is quantified as the average latency of all images on AGX Xavier.}
    \label{fig:nmodels}
\end{figure}

The number of models cascaded for dynamic inference is also crucial to the final performance of our framework. We perform comprehensive experiments to identify the optimal number of models from the perspective of accuracy, computational complexity, and actual inference latency. The experimental results in Fig. \ref{fig:nmodels} uncover an interesting insight that using too many models can worsen the trade-off between model efficiency and accuracy. 
\major{Specifically, we observe that using two models for dynamic inference achieves the optimal trade-off between model efficiency and accuracy among all configurations. Meanwhile, the two-model configuration avoids loading too many models onto the device, optimizing the memory occupation of dynamic inference.}


\subsection{Ablation Study}

Our framework contains three novel components: 1) Dynamic Image Cropping (DIC), 2) Compound Shrinking (CS), and 3) Dynamic Inference (DI). To validate the efficacy and efficiency of each component separately, we conduct ablation experiments on the ImageNet-1K dataset. The experimental results are shown in TABLE \ref{tab:ablation}, where we observe that our DIC framework (i.e., EC-DIC) outperforms the ResizedCenterCrop strategy by a remarkable 1.5\% accuracy improvement. Afterwards, we gradually integrate the CS module and DI strategy with DIC to build EC-Static and EC-Dynamic. The experimental results demonstrate that both EC-Static and EC-Dynamic further improve the accuracy, and the complete framework EC-Dynamic achieves the best accuracy. Thanks to the novel design of our framework, a significant improvement on accuracy is achieved at a slightly higher latency cost, optimizing the trade-off between accuracy and execution efficiency. The ablation experiments reveal that all DIC, CS, and DI components contribute to the final performance.

\begin{figure}
    \centering
    \includegraphics[width=0.47\textwidth]{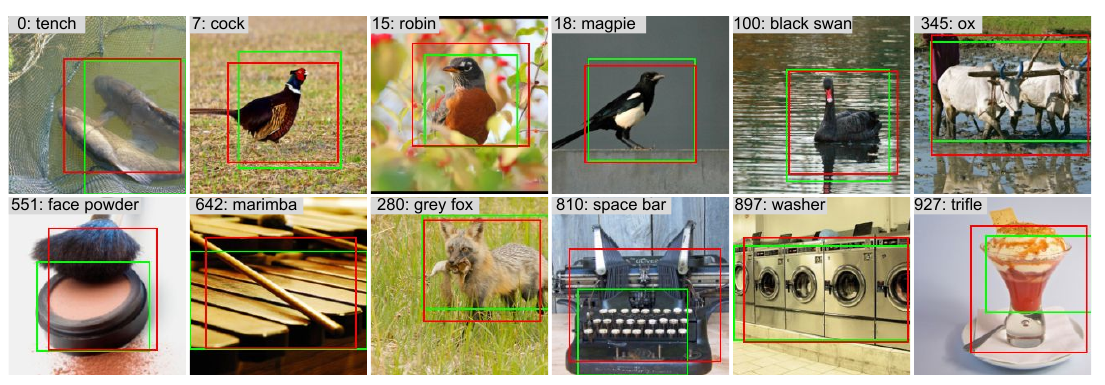}
    \caption{Visualization of the predicted bounding boxes (red) and the ground truth bounding boxes generated from Grad-CAM (green). Our predictor achieves a high localization accuracy of 62.1\% mAP on ImageNet-1K validation set. The images above are randomly selected from ImageNet-1K.}
    \label{fig:visualbox}
\end{figure}



\subsection{Visualization}

\subsubsection{Foreground Prediction}
We visualize the bounding boxes generated from both Grad-CAM and our predictor in Fig. \ref{fig:visualbox}. We can see that the foreground of most images only occupies part of the whole images, thus performing inference on the whole image is unnecessary and inefficient, which coincides with our motivation. \major{Moreover, Fig. \ref{fig:visualbox} validates that, even though the lightweight foreground predictor only has very limited computation and parameters, it can still accurately and efficiently localize the main object.
Due to the design of the efficient foreground predictor, we can remove the spatial redundancy in images, accelerating CNNs on edge devices.}

\subsubsection{Easy Samples \& Hard Samples}

\begin{figure}[!tbp]
    \centering
    \includegraphics[width=0.47\textwidth]{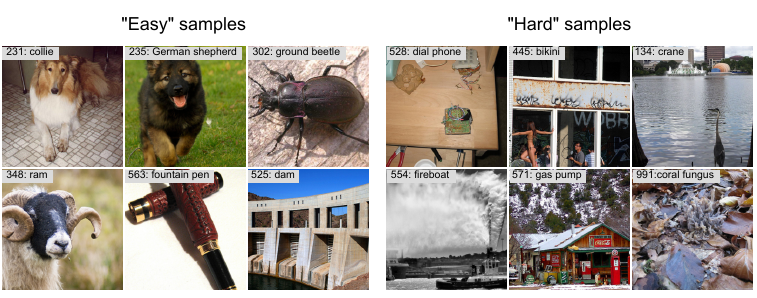}
    \caption{Visualization of some hard samples and easy samples. Hard samples are considered as the images that cannot be confidently classified by the first model, and easy samples refer to those images that can be confidently classified by the first model. All images are from ImageNet-1K.}
    \label{fig:hard_easy}
\end{figure}

\begin{table}
  \centering
  \caption{Results of ablation experiments. The baseline network is ResNet-50 and the target dataset is ImageNet-1K.}
    \resizebox{0.47\textwidth}{!}{
    \begin{tabular}{lrrr}
    \toprule
    \toprule
    \textbf{Method} & \textbf{\#MACs (B)} & \textbf{Latency (ms)} & \textbf{Top1 Acc. (\%)} \\
    \midrule
    ResNet-50 & 4.1  & 10.6 & 76.0 \\
    \midrule
    ResizedCenterCrop (RCC) & 1.8  & \textbf{6.3}  & 73.4 \\
    EC-DIC (DIC only) & 1.9  & 7.3  & 74.9 \\
    EC-Static (DIC + CS) & 1.8  & 8.3  & 75.6 \\
    EC-Dynamic (DIC + CS + DI) & \textbf{1.8}  & 7.0  & \textbf{75.9} \\
    \bottomrule
    \bottomrule
    \end{tabular}%
    }
  \label{tab:ablation}%
\end{table}%

\major{We visualize some easy samples and hard samples from ImageNet to more intuitively demonstrate the difference between them. The visualization results in \ref{fig:hard_easy} indicate that the most of easy images have a simple and clear foreground, and thus they can be correctly recognized by a small model. For hard samples, the images are more confusing because of their unintuitive foreground, and larger models are needed to mine high-level semantics in images for the correct prediction. Through the proposed dynamic inference framework, images with different difficulties can be processed by the appropriate model, achieving higher run-time efficiency and accuracy.}

\section{Conclusion}
\label{sec:conlusion}


In this paper, we propose \edge, a comprehensive CNN compression framework to reduce the computational redundancy in both input images and network architectures, facilitating the deployment of advanced CNNs onto embedded devices. In \edge, we first introduce dynamic image cropping, which effectively and efficiently removes the redundancy in input images. Subsequently, we present compound shrinking to collaboratively compress the three dimensions of CNNs, reducing the computational redundancy in both input images and network architectures. Finally, we design a dynamic inference strategy, which adaptively execute different models for different input images, further improving the inference efficiency of CNNs. Extensive experiments validate the advantages of \edge over existing SOTA approaches.
\section{Acknowledgement}
This study is partially supported under the RIE2020 Industry Alignment Fund – Industry Collaboration Projects (IAF-ICP) Funding Initiative, as well as cash and in-kind contribution from the industry partner, HP Inc., through the HP-NTU Digital Manufacturing Corporate Lab (I1801E0028). This work is also partially supported by the Ministry of Education, Singapore, under its Academic Research Fund Tier 2 (MOE2019-T2-1-071), and Nanyang Technological University, Singapore, under its NAP (M4082282).

\bibliographystyle{IEEEtran}
\bibliography{reference}

\begin{thebibliography}{10}
\providecommand{\url}[1]{#1}
\csname url@samestyle\endcsname
\providecommand{\newblock}{\relax}
\providecommand{\bibinfo}[2]{#2}
\providecommand{\BIBentrySTDinterwordspacing}{\spaceskip=0pt\relax}
\providecommand{\BIBentryALTinterwordstretchfactor}{4}
\providecommand{\BIBentryALTinterwordspacing}{\spaceskip=\fontdimen2\font plus
\BIBentryALTinterwordstretchfactor\fontdimen3\font minus
  \fontdimen4\font\relax}
\providecommand{\BIBforeignlanguage}[2]{{%
\expandafter\ifx\csname l@#1\endcsname\relax
\typeout{** WARNING: IEEEtran.bst: No hyphenation pattern has been}%
\typeout{** loaded for the language `#1'. Using the pattern for}%
\typeout{** the default language instead.}%
\else
\language=\csname l@#1\endcsname
\fi
#2}}
\providecommand{\BIBdecl}{\relax}
\BIBdecl

\bibitem{deng2009imagenet}
J.~Deng \emph{et~al.}, ``Imagenet: A large-scale hierarchical image database,''
  in \emph{CVPR}, 2009.

\bibitem{lin2014microsoft}
T.-Y. Lin \emph{et~al.}, ``Microsoft coco: Common objects in context,'' in
  \emph{ECCV}, 2014.

\bibitem{tan2019efficientnet}
M.~Tan and Q.~Le, ``Efficientnet: Rethinking model scaling for convolutional
  neural networks,'' in \emph{ICML}, 2019.

\bibitem{radosavovic2020designing}
I.~Radosavovic \emph{et~al.}, ``Designing network design spaces,'' in
  \emph{CVPR}, 2020.

\bibitem{liu2022convnet}
Z.~Liu \emph{et~al.}, ``A convnet for the 2020s,'' in \emph{CVPR}, 2022.

\bibitem{shi2016edge}
W.~Shi \emph{et~al.}, ``Edge computing: Vision and challenges,'' \emph{IoT
  Journal}, 2016.

\bibitem{Chen_2022_CVPR}
B.~\color{black} Chen \emph{et~al.}, ``Update compression for deep neural
  networks on the edge,'' in \emph{CVPR Workshops}, 2022\color{black}.

\bibitem{sandler2018mobilenetv2}
M.~Sandler \emph{et~al.}, ``Mobilenetv2: Inverted residuals and linear
  bottlenecks,'' in \emph{CVPR}, 2018.

\bibitem{tan2019mnasnet}
M.~Tan \emph{et~al.}, ``Mnasnet: Platform-aware neural architecture search for
  mobile,'' in \emph{CVPR}, 2019.

\bibitem{zhu2021dynamic}
M.~Zhu \emph{et~al.}, ``Dynamic resolution network,'' in \emph{NeurIPS}, 2021.

\bibitem{lin2020hrank}
M.~Lin \emph{et~al.}, ``Hrank: Filter pruning using high-rank feature map,'' in
  \emph{CVPR}, 2020.

\bibitem{wang2021convolutional}
Z.~Wang \emph{et~al.}, ``Convolutional neural network pruning with structural
  redundancy reduction,'' in \emph{CVPR}, 2021.

\bibitem{yu2019slimmable}
J.~Yu \emph{et~al.}, ``Slimmable neural networks,'' in \emph{ICLR}, 2019.

\bibitem{molchanov2019importance}
P.~Molchanov \emph{et~al.}, ``Importance estimation for neural network
  pruning,'' in \emph{CVPR}, 2019.

\bibitem{wang2019dbp}
W.~Wang \emph{et~al.}, ``Dbp: Discrimination based block-level pruning for deep
  model acceleration,'' \emph{arXiv preprint arXiv:1912.10178}, 2019.

\bibitem{alwani2022decore}
M.~Alwani \emph{et~al.}, ``Decore: Deep compression with reinforcement
  learning,'' in \emph{CVPR}, 2022.

\bibitem{wang2020glance}
Y.~Wang \emph{et~al.}, ``Glance and focus: A dynamic approach to reducing
  spatial redundancy in image classification,'' in \emph{NeurIPS}, 2020.

\bibitem{bochkovskiy2020yolov4}
A.~Bochkovskiy \emph{et~al.}, ``Yolov4: Optimal speed and accuracy of object
  detection,'' \emph{arXiv preprint arXiv:2004.10934}, 2020.

\bibitem{liu2016ssd}
W.~Liu \emph{et~al.}, ``Ssd: Single shot multibox detector,'' in \emph{ECCV},
  2016.

\bibitem{ren2015faster}
S.~Ren \emph{et~al.}, ``Faster r-cnn: Towards real-time object detection with
  region proposal networks,'' in \emph{NeurIPS}, 2015.

\bibitem{he2017mask}
K.~He \emph{et~al.}, ``Mask r-cnn,'' in \emph{ICCV}, 2017.

\bibitem{Everingham10}
M.~Everingham \emph{et~al.}, ``The pascal visual object classes (voc)
  challenge,'' \emph{IJCV}, 2010.

\bibitem{wei2019unsupervised}
X.-S. Wei \emph{et~al.}, ``Unsupervised object discovery and co-localization by
  deep descriptor transformation,'' \emph{Pattern Recognition}, 2019.

\bibitem{zhang2020rethinking}
C.-L. Zhang \emph{et~al.}, ``Rethinking the route towards weakly supervised
  object localization,'' in \emph{CVPR}, 2020.

\bibitem{zhou2016learning}
B.~Zhou \emph{et~al.}, ``Learning deep features for discriminative
  localization,'' in \emph{CVPR}, 2016.

\bibitem{selvaraju2017grad}
R.~R. Selvaraju \emph{et~al.}, ``Grad-cam: Visual explanations from deep
  networks via gradient-based localization,'' in \emph{CVPR}, 2017.

\bibitem{zhang2018adversarial}
X.~Zhang \emph{et~al.}, ``Adversarial complementary learning for weakly
  supervised object localization,'' in \emph{CVPR}, 2018.

\bibitem{liu2022brining}
D.~Liu \emph{et~al.}, ``Bringing ai to edge: From deep learning’s
  perspective,'' \emph{Neurocomputing}, 2022.

\bibitem{zhang2022layer}
K.~Zhang and G.~Liu, ``Layer pruning for obtaining shallower resnets,''
  \emph{IEEE Signal Processing Letters}, 2022.

\bibitem{teerapittayanon2016branchynet}
S.~Teerapittayanon \emph{et~al.}, ``Branchynet: Fast inference via early
  exiting from deep neural networks,'' in \emph{ICPR}, 2016.

\bibitem{huang2018multiscale}
G.~Huang \emph{et~al.}, ``Multi-scale dense networks for resource efficient
  image classification,'' in \emph{ICLR}, 2018.

\bibitem{bolukbasi2017adaptive}
T.~Bolukbasi \emph{et~al.}, ``Adaptive neural networks for efficient
  inference,'' in \emph{ICML}, 2017.

\bibitem{graves2016adaptive}
A.~Graves, ``Adaptive computation time for recurrent neural networks,''
  \emph{arXiv preprint arXiv:1603.08983}, 2016.

\bibitem{wang2018skipnet}
X.~Wang \emph{et~al.}, ``Skipnet: Learning dynamic routing in convolutional
  networks,'' in \emph{ECCV}, 2018.

\bibitem{veit2018convolutional}
A.~Veit and S.~Belongie, ``Convolutional networks with adaptive inference
  graphs,'' in \emph{ECCV}, 2018.

\bibitem{hua2019channel}
W.~Hua \emph{et~al.}, ``Channel gating neural networks,'' \emph{NeurIPS}, 2019.

\bibitem{lin2017runtime}
J.~Lin \emph{et~al.}, ``Runtime neural pruning,'' \emph{NeurIPS}, 2017.

\bibitem{gao2018dynamic}
X.~Gao \emph{et~al.}, ``Dynamic channel pruning: Feature boosting and
  suppression,'' in \emph{ICLR}, 2019.

\bibitem{herrmann2020channel}
C.~Herrmann \emph{et~al.}, ``Channel selection using gumbel softmax,'' in
  \emph{ECCV}, 2020.

\bibitem{yang2020resolution}
L.~Yang \emph{et~al.}, ``Resolution adaptive networks for efficient
  inference,'' in \emph{CVPR}, 2020.

\bibitem{he2016deep}
K.~He \emph{et~al.}, ``Deep residual learning for image recognition,'' in
  \emph{CVPR}, 2016.

\bibitem{li2018tiny}
Y.~Li \emph{et~al.}, ``Tiny-dsod: Lightweight object detection for
  resource-restricted usages,'' in \emph{BMVC}, 2018.

\bibitem{kingma2015adam}
D.~P. Kingma and J.~Ba, ``Adam: A method for stochastic optimization,'' in
  \emph{ICLR}, 2015.

\bibitem{Li2020An}
Z.~Li and S.~Arora, ``An exponential learning rate schedule for deep
  learning,'' in \emph{ICLR}, 2020.

\bibitem{laskaridis2020hapi}
S.~Laskaridis \emph{et~al.}, ``Hapi: Hardware-aware progressive inference,'' in
  \emph{ICCAD}, 2020.

\bibitem{kaya2019shallow}
Y.~Kaya \emph{et~al.}, ``Shallow-deep networks: Understanding and mitigating
  network overthinking,'' in \emph{ICML}, 2019.

\bibitem{simonyan2014very}
K.~\color{black}Simonyan and A.~Zisserman, ``Very deep convolutional networks
  for large-scale image recognition,'' \emph{arXiv preprint arXiv:1409.1556},
  2014\color{black}.

\bibitem{szegedy2016rethinking}
C.~Szegedy \emph{et~al.}, ``Rethinking the inception architecture for computer
  vision,'' in \emph{CVPR}, 2016.

\bibitem{zagoruyko2016wide}
S.~Zagoruyko and N.~Komodakis, ``Wide residual networks,'' in \emph{BMVC},
  2016.

\bibitem{huang2018data}
Z.~Huang and N.~Wang, ``Data-driven sparse structure selection for deep neural
  networks,'' in \emph{ECCV}, 2018.

\bibitem{wang2018learning}
Y.~Wang \emph{et~al.}, ``Learning versatile filters for efficient convolutional
  neural networks,'' in \emph{NeurIPS}, 2018.

\bibitem{liebenwein2019provable}
L.~Liebenwein \emph{et~al.}, ``Provable filter pruning for efficient neural
  networks,'' in \emph{ICLR}, 2019.

\bibitem{ding2019centripetal}
X.~Ding \emph{et~al.}, ``Centripetal sgd for pruning very deep convolutional
  networks with complicated structure,'' in \emph{CVPR}, 2019.

\bibitem{lin2019towards}
S.~Lin \emph{et~al.}, ``Towards optimal structured cnn pruning via generative
  adversarial learning,'' in \emph{CVPR}, 2019.

\bibitem{luo2020autopruner}
J.-H. Luo and J.~Wu, ``Autopruner: An end-to-end trainable filter pruning
  method for efficient deep model inference,'' \emph{Pattern Recognition},
  2020.

\bibitem{huang2017densely}
G.~Huang \emph{et~al.}, ``Densely connected convolutional networks,'' in
  \emph{CVPR}, 2017.

\bibitem{ioffe2015batch}
S.~Ioffe and C.~Szegedy, ``Batch normalization: Accelerating deep network
  training by reducing internal covariate shift,'' in \emph{ICML}, 2015.

\bibitem{zhao2019variational}
C.~\color{black}Zhao \emph{et~al.}, ``Variational convolutional neural network
  pruning,'' in \emph{CVPR}, 2019\color{black}.

\end{thebibliography}

\begin{IEEEbiography}[{\includegraphics[width=1in,height=1.25in,clip,keepaspectratio]{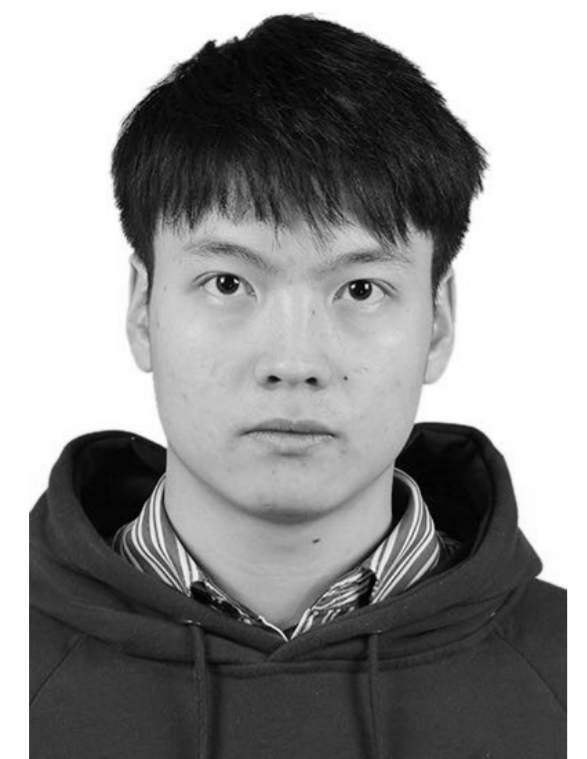}}]{Hao Kong} received the B.E. degree from University of Electronic Science and Technology of China, China, in 2019. He is currently a Ph.D. student at the School of Computer Science and Engineering, Nanyang Technological University, Singapore. His research interests include edge AI acceleration, model compression, and adaptive deep learning.
\vspace{-20pt}
\end{IEEEbiography}

\begin{IEEEbiography}[{\includegraphics[width=1in,height=1.25in,clip,keepaspectratio]{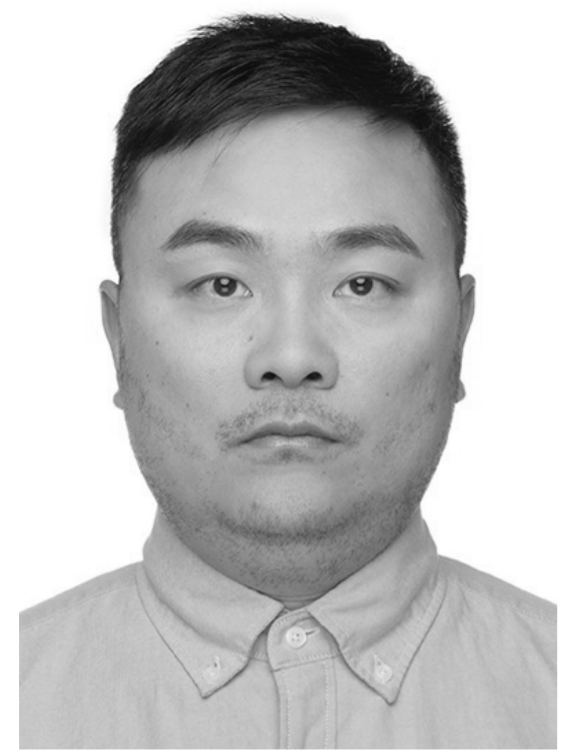}}]{Di Liu}  is an Associate Professor at Department of Computer Science, Norwegian University of Science and Technology (NTNU), Norway. In prior to that, he was a research fellow at Nanyang Technological University, Singapore and a faculty member at Yunnan University, China. He received his PhD degree from Leiden University, The Netherlands, and both his MEng and BEng degrees from Northwestern Polytechnical University, China.
\end{IEEEbiography}

\begin{IEEEbiography}
[{\includegraphics[width=1in,height=1.25in,clip,keepaspectratio]{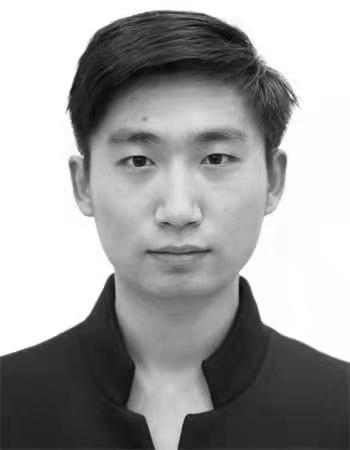}}]{Shuo Huai} received the B.E. degree from the School of Computer Science at Shandong University, China, in 2019. He is currently a Ph.D. student at the School of Computer Science and Engineering at Nanyang Technological University, Singapore. His research interests are efficient deep learning algorithms, embedded intelligence, and in-memory computing.
\end{IEEEbiography}

\begin{IEEEbiography}
[{\includegraphics[width=1in,height=1.25in,clip,keepaspectratio]{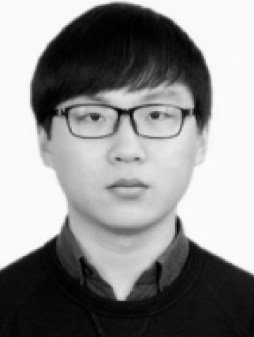}}]{Xiangzhong Luo} 
received the B.E. degree from Shanghai Jiao Tong University, Shanghai, China, in 2019. He is currently pursuing the Ph.D degree from the School of Computer Science and Engineering at Nanyang Technological University, Singapore. His current research interests include edge intelligence, hardware-aware neural architecture search, and model compression.
\end{IEEEbiography}

\begin{IEEEbiography}
[{\includegraphics[width=1in,height=1.25in,clip,keepaspectratio]{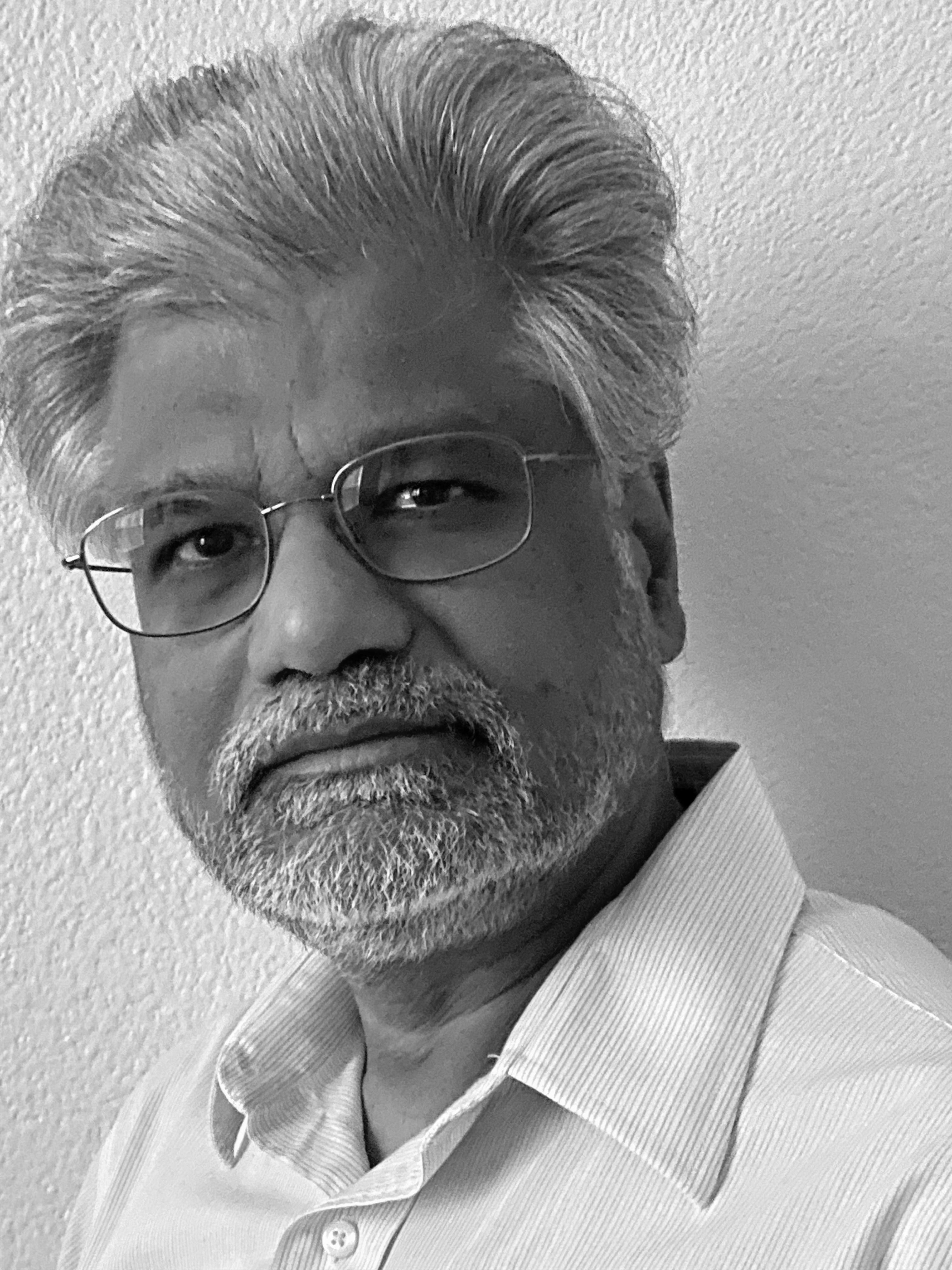}}]{Ravi Subramaniam} is a Distinguished Technologist at HP Inc working on Machine Learning algorithms and system architectures for ML at the edge and for low power and embedded systems. He has led solution development and standards work in IOT, P2P systems, distributed systems, Cloud computing, Operating systems over the last 28 years.
\end{IEEEbiography}

\begin{IEEEbiography}
[{\includegraphics[width=1in,height=1.25in,clip,keepaspectratio]{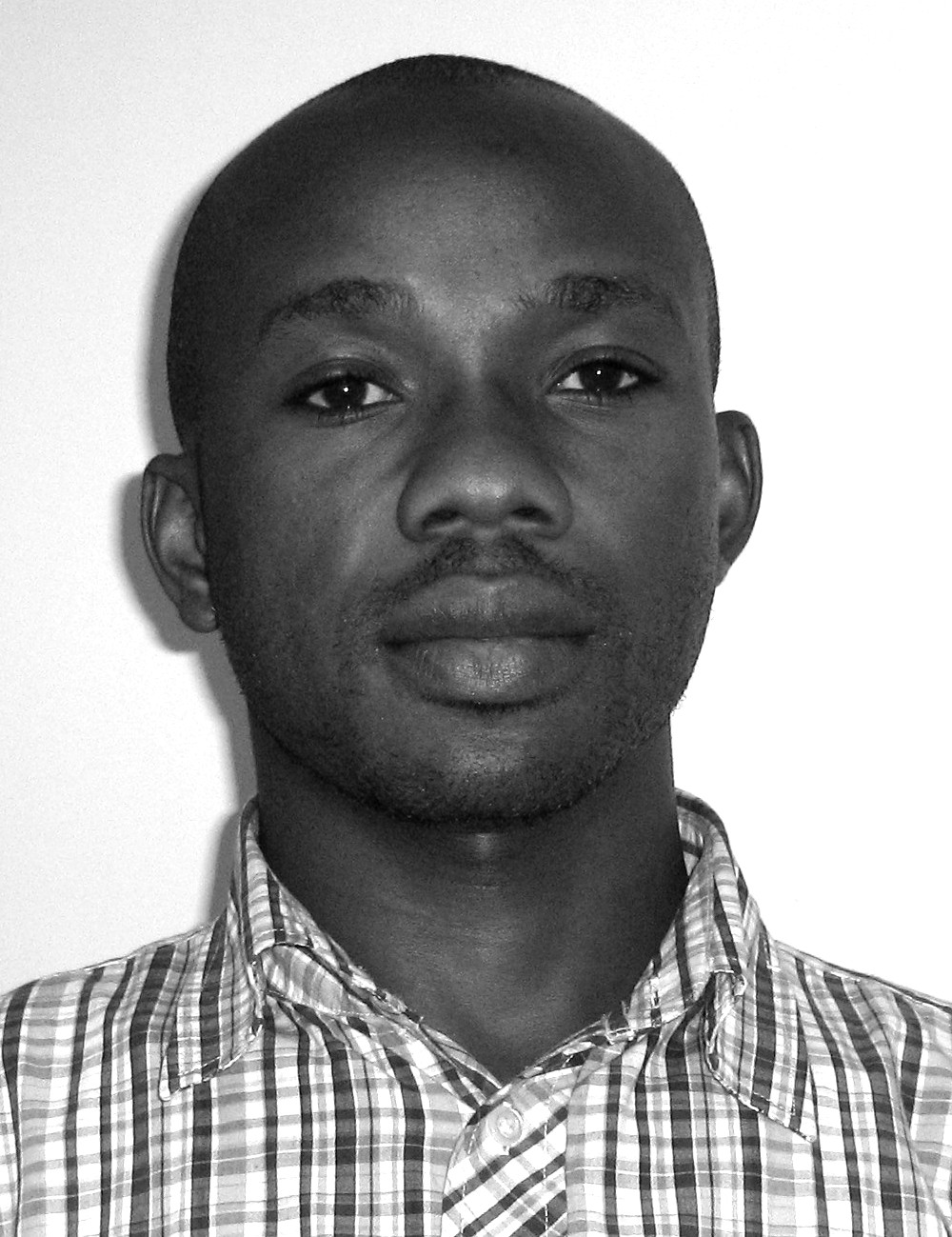}}]{Christian Makaya} is currently a Principal Research Scientist at HP Labs, HP inc., Palo Alto, CA.  He received the Ph.D. degree and the M.Sc. degree from University of Montreal, Quebec, Canada in 2007 and 2003, respectively. The focus of his current research interests is on artificial intelligence, machine learning, deep learning, distributed computing systems, edge computing, security and privacy, and network intelligence. He is a Senior Member of IEEE and serves on the Industry Outreach Board of IEEE Communication Society (ComSoc). He served as the co-chair of IEEE Young Professionals for the IEEE Princeton/Central Jersey section.
\end{IEEEbiography}

\begin{IEEEbiography}
[{\includegraphics[width=1in,height=1.25in,clip,keepaspectratio]{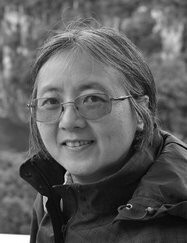}}]{Qian Lin} is a HP Fellow and Chief Technologist for Machine Learning and Vision Systems in Personal Systems Software (PSSW). She is also an adjunct professor at Purdue University, supervising PhD students in their deep learning research. Dr. Lin joined Hewlett-Packard Company in 1992. She received her BS from Xi’an Jiaotong University in China, her MSEE from Purdue University, and her Ph.D. in Electrical Engineering from Stanford University. Dr. Lin is inventor/co-inventor for 45 issued patents. She was awarded Fellowship by the Society of Imaging Science and Technology (IS\&T) in 2012, and Outstanding Electrical Engineer by the School of Electrical and Computer Engineering of Purdue University in 2013. Dr. Lin received SWE Achievement Award in 2021.
\end{IEEEbiography}

\begin{IEEEbiography}
[{\includegraphics[width=1in,height=1.25in,clip,keepaspectratio]{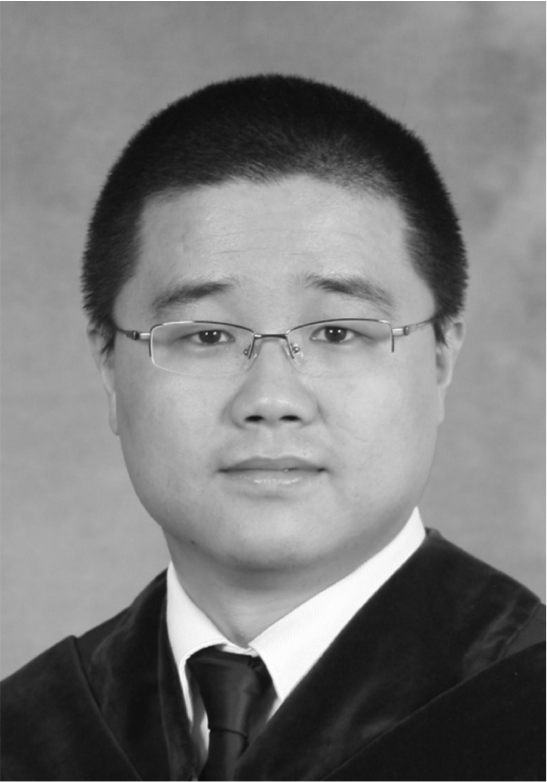}}]{Weichen Liu} is a Nanyang Assistant Professor at the School of Computer Science and Engineering, Nanyang Technological University, Singapore. He received the Ph.D. degree from the Hong Kong University of Science and Technology in 2011, and the B.Eng. and M.Eng. degrees from Harbin Institute of Technology, China, in 2004 and 2006, respectively. Dr. Liu authored and co-authored more than 100 research papers in peer-reviewed journals, conferences, and books. His research interests include embedded and real-time systems, multiprocessor systems, and machine learning accelerators.
\end{IEEEbiography}

\end{document}